\pdfoutput=1

\documentclass[11pt]{article}

\usepackage[final]{acl}

\usepackage{CJKutf8}

\usepackage{times}
\usepackage{latexsym}

\usepackage{amsmath,amsfonts,bm}

\def\eqref#1{equation~\ref{#1}}

\def\1{\bm{1}}

\def\vg{{\bm{g}}}

\def\vs{{\bm{s}}}

\def\vx{{\bm{x}}}
\def\vy{{\bm{y}}}

\DeclareMathAlphabet{\mathsfit}{\encodingdefault}{\sfdefault}{m}{sl}
\SetMathAlphabet{\mathsfit}{bold}{\encodingdefault}{\sfdefault}{bx}{n}

\usepackage[T1]{fontenc}

\usepackage[utf8]{inputenc}

\usepackage{microtype}

\usepackage{inconsolata}

\usepackage{graphicx}
\usepackage{multirow}
\usepackage{booktabs}

\newcommand{\method}{InfiniSST}

\title{\method: Simultaneous Translation of Unbounded Speech with Large Language Model}

\author{Siqi Ouyang, Xi Xu, Lei Li \\
         Carnegie Mellon University, Language Technologies Institute \\
         \texttt{\{siqiouya, xixu, leili\}@andrew.cmu.edu}}

\begin{document}
\maketitle

\begin{abstract}

Simultaneous translation of unbounded streaming speech remains a challenging problem due to the need for effectively processing the historical speech context and past translations so that quality and latency, including computation overhead, can be balanced. Most prior works assume pre-segmented speech, limiting their real-world applicability. In this paper, we propose \method, a novel approach that formulates SST as a multi-turn dialogue task, enabling seamless translation of unbounded speech. We construct translation trajectories and robust segments from MuST-C with multi-latency augmentation during training and develop a key-value (KV) cache management strategy to facilitate efficient inference. Experiments on MuST-C En-Es, En-De, and En-Zh demonstrate that \method~reduces computation-aware latency by 0.5 to 1 second while maintaining the same translation quality compared to baselines. Ablation studies further validate the contributions of our data construction and cache management strategy\footnote{\url{https://github.com/LeiLiLab/InfiniSST}}.
\end{abstract}
\section{Introduction}

Simultaneous speech translation (SST) is the task of translating partial speech input from a source language into text in a target language, with a wide range of applications, including conference interpretation and live-streaming translation~\cite{ma-etal-2020-simulmt,ren-etal-2020-simulspeech}.
Most prior research on SST focuses on translating pre-segmented speech (SST-S), assuming that ground-truth segmentation is provided~\cite{liu-etal-2021-cross,zeng-etal-2021-realtrans,dong-etal-2022-learning,alignatt,papi2024realrealtimesimultaneousspeechtotext}. However, translating unbounded, streaming speech (SST-U) remains underexplored.

Unbounded speech presents a major challenge that the model has to effectively process the historical speech context and past translations so that quality and latency, including computation overhead, can be balanced. 
Large language model (LLM) is a promising solution for long-context modeling with the recent advancements~\cite{Su2021RoFormerET,xiao2024efficientstreaminglanguagemodels,han-etal-2024-lm}. Moreover, LLM-based architectures have been shown to improve SST-S performance~\cite{xu-etal-2024-cmus}. However, conventional SST-S approaches suffer from high computational costs, as they require recomputing features for past speech and generated text every time a new speech chunk arrives. Some studies mitigate this issue by framing SST as a multi-turn dialogue task, either explicitly~\cite{yu2025simulpl,conversationsimulmt} or implicitly~\cite{ouyang2024fasstfastllmbasedsimultaneous,raffel-etal-2024-simultaneous}, leveraging key-value (KV) caching to improve efficiency. While effective for segmented speech and text, these methods do not seamlessly extend to unbounded speech.

In this paper, we propose \method, a method for simultaneous translation of unbounded speech using a multi-turn dialogue format. We construct SST trajectories and derive robust speech segments for training from the MuST-C dataset, enhancing them with a multi-latency strategy to increase diversity. During inference, we employ a KV cache management strategy, inspired by \citet{han-etal-2024-lm}, to enable seamless extrapolation to unbounded speech input. Experiments on MuST-C En-Es, En-De, and En-Zh~\cite{di-gangi-etal-2019-must} show that \method~reduces computation-aware latency by 0.5 to 1 second while maintaining the same BLEU score as baselines. A detailed ablation study further validates the effectiveness of our data construction and cache management strategies during inference.

\section{Related Works}

\subsection{SST on Unbounded Speech} 

\paragraph{Cascade Approaches} Cascade-based methods typically use an automatic speech recognition (ASR) model to segment and transcribe the input, followed by a machine translation model that translates the transcription~\cite{1660084, yoshimura2020endtoendautomaticspeechrecognition, huang2022e2esegmenterjointsegmenting, donato-etal-2021-diverse,iranzo-sanchez-etal-2024-segmentation}. However, segmentation errors and the lack of punctuation degrade translation quality, which complicates maintaining low latency and high quality.

\paragraph{Direct SST on Unbounded Speech} 
Several works explore end-to-end approaches for SST on unbounded speech~\cite{schneider-waibel-2020-towards,papi-etal-2024-streamatt}. These methods avoid external segmentation by dynamically preserving relevant audio context and previously generated text while discarding older information. \citet{papi-etal-2024-streamatt} extends AlignAtt to unbounded speech by storing text and audio history in a fully streaming way, which helps reduce latency and maintain contextual awareness. Despite these advances, balancing translation quality, latency, and computational demands remains a challenge. Our approach addresses these issues by managing unbounded speech input without loss in translation accuracy and with improved computational efficiency.

\subsection{Length Extrapolation of LLM}

Recent advances in positional encoding~\cite{Su2021RoFormerET,Press2021TrainST,sun-etal-2023-length} have enabled models to handle longer sequences with little or no additional training.
ReRoPE~\cite{rerope2023} introduces an NTK-aware Scaled RoPE that extends context length to infinite without fine-tuning. \citet{han-etal-2024-lm} and \citet{xiao2024efficientstreaminglanguagemodels} propose on-the-fly length generalization based on a $\boldsymbol{\Lambda}$-shaped attention window, allowing nearly unlimited input length with no fine-tuning.
\method~is a successful application of RoPE and $\boldsymbol{\Lambda}$-shaped attention window in SST-U.


\section{Method}
\label{sec:method}

\subsection{Problem Formulation}

Let \(\vs_{1:t} = (s_1, s_2, \dots, s_t)\) be the partial input of an unbounded input speech sequence and \(\vy_{1:i} = (y_1, y_2, \dots, y_i)\) represent the partial text translation. 
Here $\vs_{1:t}$ is the waveform with a specific sampling rate. 
Define \(\pi(\vs_{1:t}, \vy_{1:i}) \in [0, 1]\) as the policy to determine whether to take more speech input (=0) or to generate target translation tokens (=1). 
Whenever $\pi(\vs_{1:t}, \vy_{1:i})=1$, we define $g_{i+1} = t$ as the delay of $i+1$-th token. Let $g_0=0$.
In addition, let $\theta$ be the model parameter, we define the probability of generating next token given a partial speech input as 
$P_{\theta}(y_{i+1} \mid \vs_{1:t}, \vy_{1:i})$.
In our formulation, we use a simple policy by checking whether the current generated token $y_i$ is a special ending token $T_0$ (e.g. stop writing translation and read speech input when encountering ``$\langle$\texttt{EOT}$\rangle$'' token in Llama~\cite{llama3}).
\begin{align}
    \pi(\vs_{1:t}, \vy_{1:i}) = 
    \begin{cases}
			0, & \text{if $y_i=T_0$}\\
            1, & \text{otherwise}
    \end{cases}
\end{align}

Given $\vs$, we define the conditional probability of generating a translation sequence $\vy_{1:i}$ with associated delays for each token $\vg_{1:i}$ as:
\begin{align}
    P(\vy,\vg|\vs) &= \prod_{i=1}^{|\vy|} \bigg(P_\theta(y_i|\vs_{1:g_i},\vy_{1:i-1}) \\    
    \pi(\vs_{1:g_i},&\vy_{1:i-1})\prod_{j=g_{i-1}}^{g_i-1}\big(1-\pi(\vs_{1:j},\vy_{i-1})\big)\bigg) \nonumber
\end{align}

The translation quality and latency are subsequently evaluated based on $\vs$, \(\vy\) and \(\vg\).

\subsection{Model Architecture}
\label{sec:arch}

\begin{figure*}[htb]
    \centering
    \includegraphics[width=\linewidth]{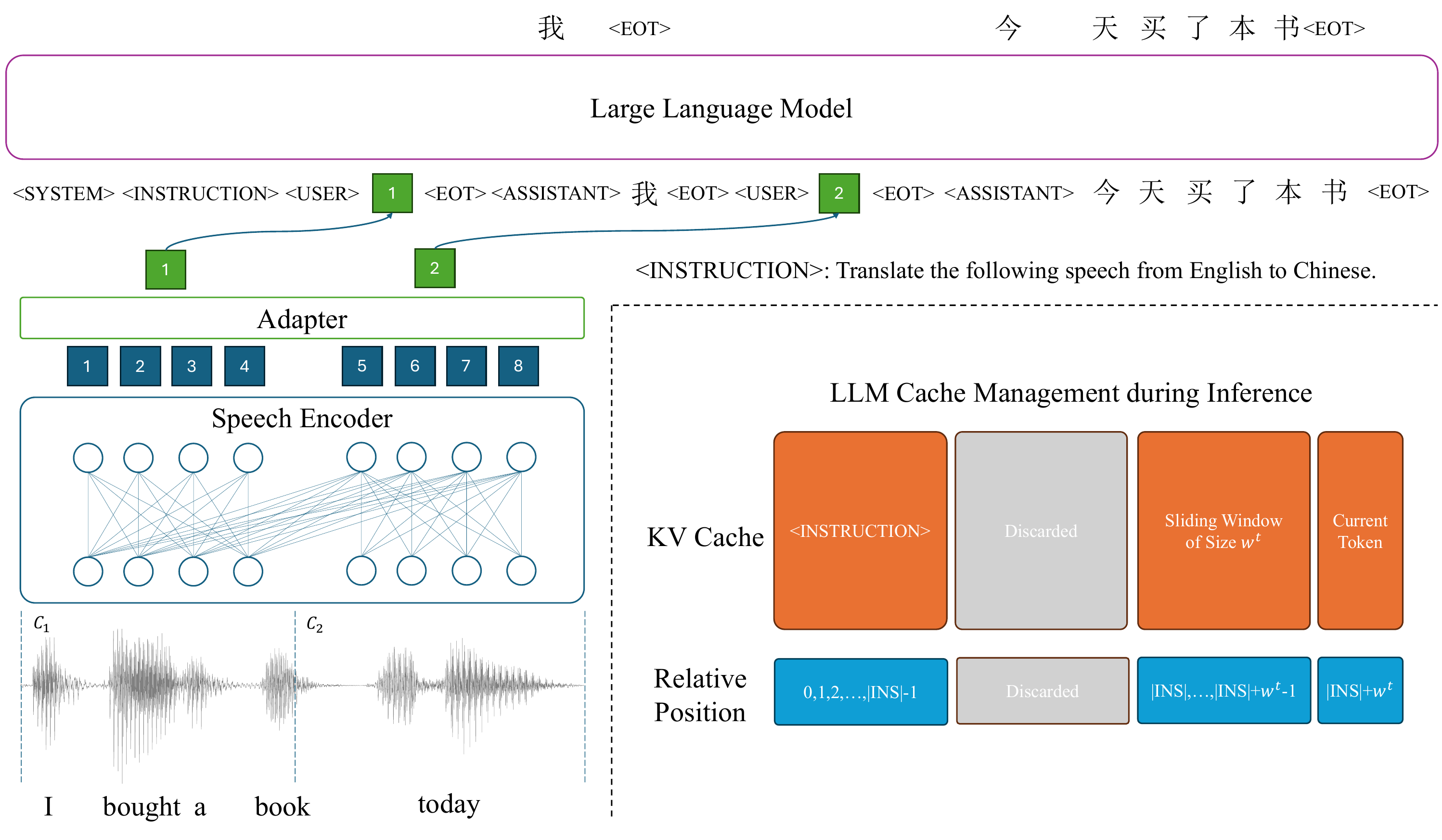}
    \caption{Model architecture of \method. \method~first encodes speech using a chunkwise-causal speech encoder, then compresses the speech features into embeddings via an adapter. The large language model (LLM) processes the input by first reading a system instruction, then alternating between consuming speech embeddings and generating translations. The translation process stops when the LLM generates an EOT token. During inference, we employ a sliding window of size $w^t$ for the LLM, conditioning the translation on the most recent $w^t$ KV caches along with the KV cache of the system instruction, enabling extrapolation to unbounded speech input.}
    \label{fig:InfiniSST-model}
\end{figure*}

We design \method, a simultaneous speech translation model that can take unbounded streaming speech input and generate target text efficiently. 
The \method~consists of 1) a streaming speech encoder to incrementally compute representations of partial speech input without recomputation, 2) a speech-to-token embedding adapter to match speech representations to LLM's token embedding space, and 3) an multi-turn LLM decoder to interactively take speech input and generate translation as needed (Figure ~\ref{fig:InfiniSST-model}).  

\paragraph{Streaming Speech Encoder} 
We modify a pre-trained wav2vec2~\cite{wav2vec2}\footnote{Wav2Vec 2.0 Large (LV-60) + Self Training on LibriSpeech. \url{https://dl.fbaipublicfiles.com/fairseq/wav2vec/wav2vec_vox_960h_pl.pt}} speech encoder to encode the unbounded streaming speech input. 
However, there is a major limitation of the original wav2vec2. 
It uses bidirectional attention and bidirectional convolutional position embedding, which needs to recompute the representations for every new segment of streaming speech input. 
To handle unbounded speech input, we introduce three modifications to the speech encoder. 
Firstly, we replace the wav2vec2's convolutional positional embedding with a rotary positional embedding (RoPE)~\cite{rope} because it shows better extensibility for long sequences. 
Secondly, we replace bidirectional attention with chunk-wise causal attention~\cite{blockcausal}. 
Each chunk contains 48 frames in wav2vec2, with a total duration of 960ms. 
The multihead attention within each chunk remains bidirectional while attention across chunks is causal. This is achieved by adding block-wise masking to the attention weights. 
Thirdly, we apply a sliding window mechanism with window size \(w^s\) to maintain a finite context length, restricting chunk \(i\) to attend only to hidden states of chunks \([i - w^s + 1, i]\).
In practice, we use $w^s=10$ so each speech embedding is computed from roughly 9.6 seconds of the preceding speech input. 

\paragraph{Speech-to-Token Embedding Adapter} 

The speech encoder yields a sequence slightly longer than the transcript and with embeddings that differ from the LLM’s token embeddings. Following~\citet{lst}, we downsample the encoder output with two 1‑D convolutions (kernel size 2, stride 2, no padding) and linearly project the result into the LLM embedding space, so 48 input frames produce 12 embedding vectors.

\paragraph{Multi-turn LLM Decoder} 
Our decoder needs to produce target text and a special token to indicate the switching from generation to taking speech input. 
To this end, we use Llama-3.1-8B-Instruct~\cite{llama3}\footnote{\url{https://huggingface.co/meta-llama/Llama-3.1-8B-Instruct}} and employ a multi-turn dialogue format to formulate the input. 
We first feed a system instruction 

\begin{center}
    \verb|Translate the following speech| \\
    \verb|from <LangX> to <LangY>.|
\end{center}

We then add a special \verb|USER| token to indicate that the following 12 embeddings and a trailing \verb|EOT| (\verb|End-Of-Turn|) token are for speech input.
We then prompt the LLM with a special \verb|ASSISTANT| token to force LLM to generate tokens. 
We add a policy module to check generated tokens. 
When the policy module encounters the special \verb|EOT| token, it will feed a special \verb|USER| token and take 12 new streaming speech embeddings with a trailing \verb|EOT| token as new input to the LLM. 
We will describe later our inference method to incrementally compute embeddings and generate target tokens for infinite speech input.

\subsection{Training Data Construction}
\label{sec:data}

\paragraph{SST Trajectory} 
Common speech translation datasets like MuST-C are segmented from complete talks~\cite{di-gangi-etal-2019-must}. 
To train an SST model in a multi-turn dialogue format, we transform segmented ST triplets (speech $\vs$, transcript $\vx$, translation $\vy$) from MuST-C dataset into SST trajectories. 
An SST trajectory represents an alternating action sequence of speech reading and translation writing.

\begin{figure}
    \centering
    \includegraphics[width=\linewidth]{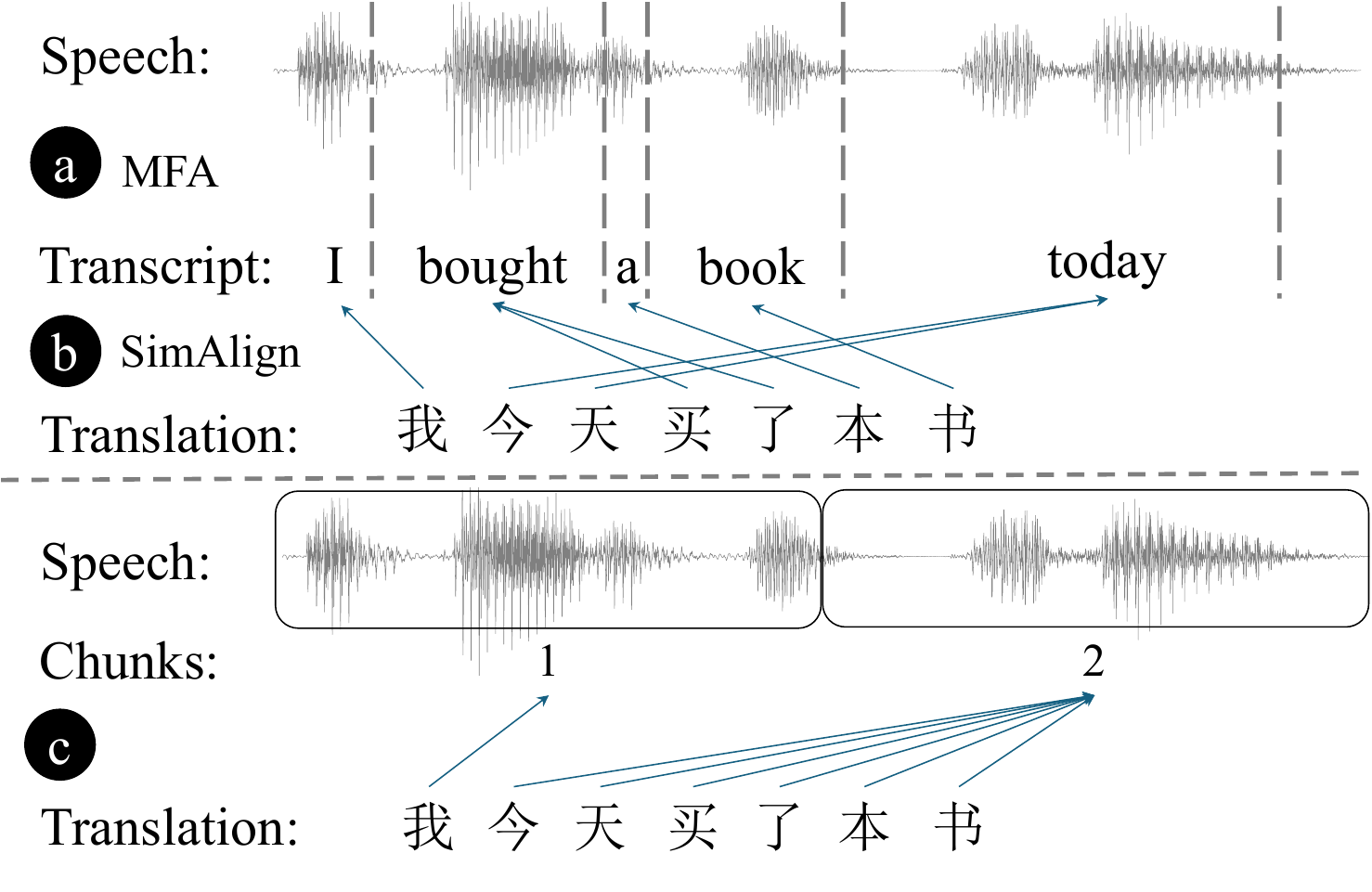}
    \caption{Trajectory construction process. We first align speech to its transcript using forced alignment, then align the transcript to its translation using SimAlign. The resulting alignments are monotonized and grouped by speech chunks.}
    \label{fig:alignment}
\end{figure}

As shown in Figure \ref{fig:alignment}, we first align speech utterances with their corresponding transcripts using the Montreal Forced Aligner (MFA)~\cite{mcauliffe17_interspeech}\footnote{\url{https://github.com/MontrealCorpusTools/Montreal-Forced-Aligner}}. Let \(m^{sx}_k\) denote the right boundary of the speech segment corresponding to the transcript token \(x_k\). Also, we utilize SimAlign~\cite{jalili-sabet-etal-2020-simalign} with the LaBSE model~\cite{feng-etal-2022-language} to align words between the transcript and translation. 
We then monotonize these alignments following \citet{conversationsimulmt}. Let \(x_{m^{xy}_i}\) be the transcript token that corresponds to translation token \(y_i\). By combining \(m^{sx}\) and \(m^{xy}\), we establish a mapping from translation token \(y_i\) to its speech boundary \(m^{sy}_i = m^{sx}_{m^{xy}_i}\) , meaning that \(y_i\) is generated after reading \(\vs_{1:m^{sy}_i}\). 

Finally, we cut the speech utterance into fixed-length chunks, each lasting 960 ms. We then concatenate translation tokens whose corresponding speech boundaries fall within the same chunk, forming a sequence of trajectory \((\vs_{C_1}, \vy_{C_1}), (\vs_{C_2}, \vy_{C_2}), \dots\).

\paragraph{Robust Segments for Training} 
Segmented speech utterances primarily consist of human speech; however, non-linguistic sounds (e.g., laughter, applause) are also present. To enhance the robustness of the SST dataset, we cut the entire talk evenly into robust segments that each span 30 speech chunks.
If a robust segment starts in the middle of a segmented speech utterance, we shift the robust segment to start with this utterance.
The trajectories for a robust segment can then be built by concatenating the trajectories of segmented utterances within this robust segment according to their timestamps and filling the rest translation entries of the trajectory as empty strings. 

\paragraph{Multi-Latency Augmentation} To further enhance trajectory diversity during training, we propose a simple yet effective multi-latency augmentation strategy. Specifically, given a trajectory \((\vs_{C_1}, \vy_{C_1}), (\vs_{C_2}, \vy_{C_2}), \dots\), we randomly select a latency multiplier \( m \in [1, M] \) and merge every \( m \) consecutive chunks of speech with their corresponding translations. The \( i \)-th step in the augmented trajectory is then represented as 
\[
(\vs_{C_{im}, \dots, C_{(i+1)m-1}}, \vy_{C_{im}, \dots, C_{(i+1)m-1}}).
\]
We also multiply the chunk size of speech encoder with $m$, i.e., number of frame in a chunk becomes $48m$.

\subsection{Training}

We train \method~with standard cross-entropy loss on translation tokens, including \verb|EOT|, of the augmented trajectory from robust segments. 
In the first stage, we freeze the LLM and train only the speech encoder and adapter. In the second stage, we freeze the speech encoder and adapter, training only the LLM.

\subsection{Inference on Unbounded Speech}
\label{sec:infer}

During inference, we cut the unbounded input speech into 960 ms chunks. The latency multiplier \( m \) during inference regulates latency by ensuring that translation begins only after every \( m \) new chunks have arrived.

At the \( i \)-th step, suppose the newly received speech chunks are \( C_{im}, \dots, C_{(i+1)m-1} \). Both the speech encoder and the LLM maintain a key-value (KV) cache to prevent redundant computations. Notably, the stored key and value features are extracted \textit{before} applying RoPE, ensuring that no positional information is embedded within the KV cache. 

The speech encoder processes the \( m \) new chunks into \( 48m \) speech features, utilizing the KV cache from chunks \( C_{im-w^s+1}, \dots, C_{im-1} \), where \( w^s \) is the sliding window size defined in Section~\ref{sec:arch}. The adapter then downsamples the \( 48m \) features into \( 12m \) embeddings, which are passed to the LLM.

As shown in Figure \ref{fig:InfiniSST-model}, the LLM employs a sliding window of size \( w^t \). By default, $w^t=1000$. Inspired by \citet{han-etal-2024-lm} and \citet{xiao2024efficientstreaminglanguagemodels}, we concatenate the KV cache of instruction with those of the most recent \( w^t \) tokens and apply RoPE on top of them. Then the LLM generate translations conditioned on this combined KV cache.

\section{Experiment Setups}

\begin{figure*}[ht]
    \centering
    \includegraphics[width=\linewidth]{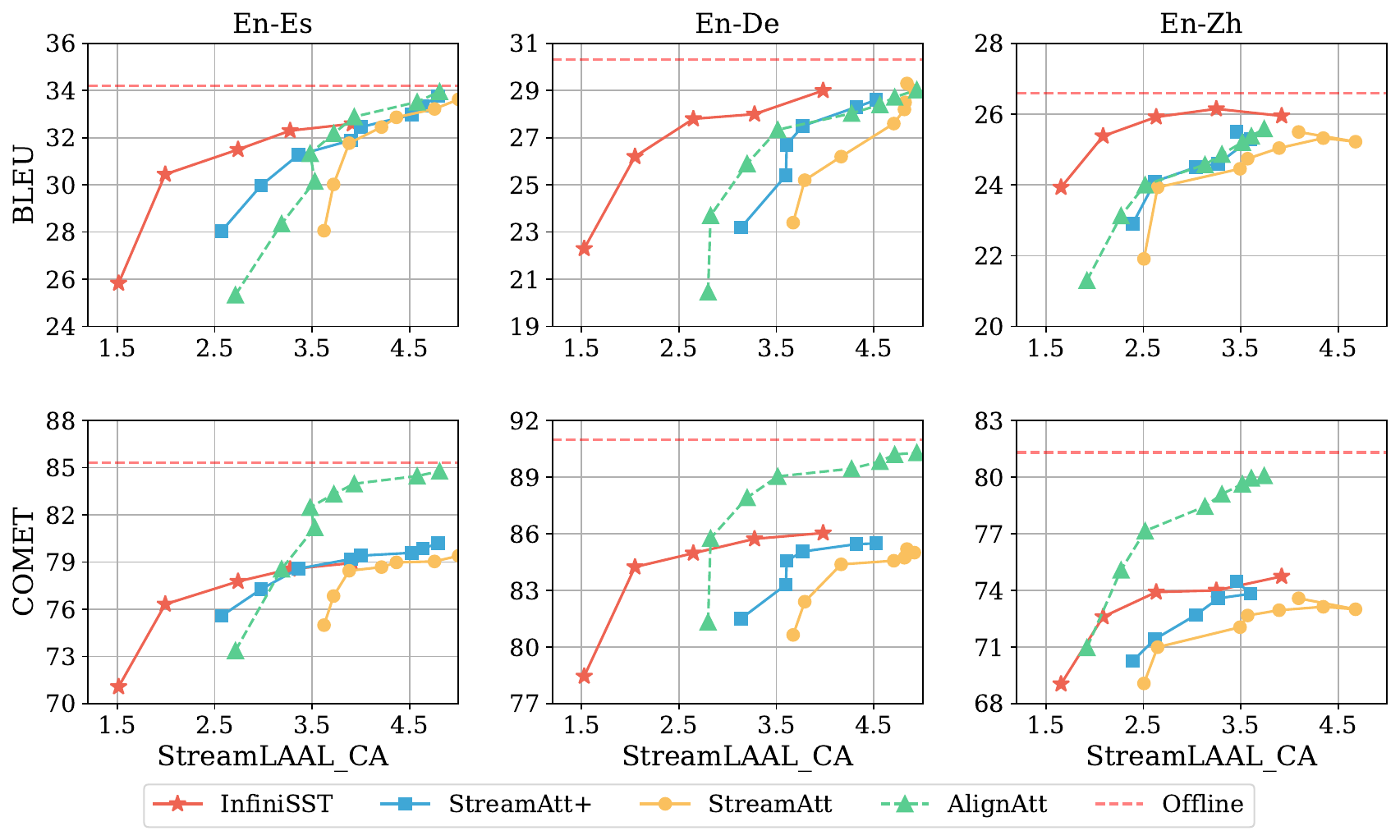}
    \caption{
    Quality-latency trade-off of \method~compared to the baselines on complete TED talks from the MuST-C tst-COMMON dataset in the En-Es, En-De, and En-Zh directions. Translation quality is measured using BLEU and COMET scores, while latency is evaluated using the \textit{computation-aware} StreamLAAL metric. For reference, we also include offline translation quality and results from AlignAtt tested on segmented speech. \method~achieves significantly lower computation-aware latency compared to StreamAtt at the same quality.}
    \label{fig:main_ca}
\end{figure*}

\begin{figure*}[ht]
    \centering
    \includegraphics[width=\linewidth]{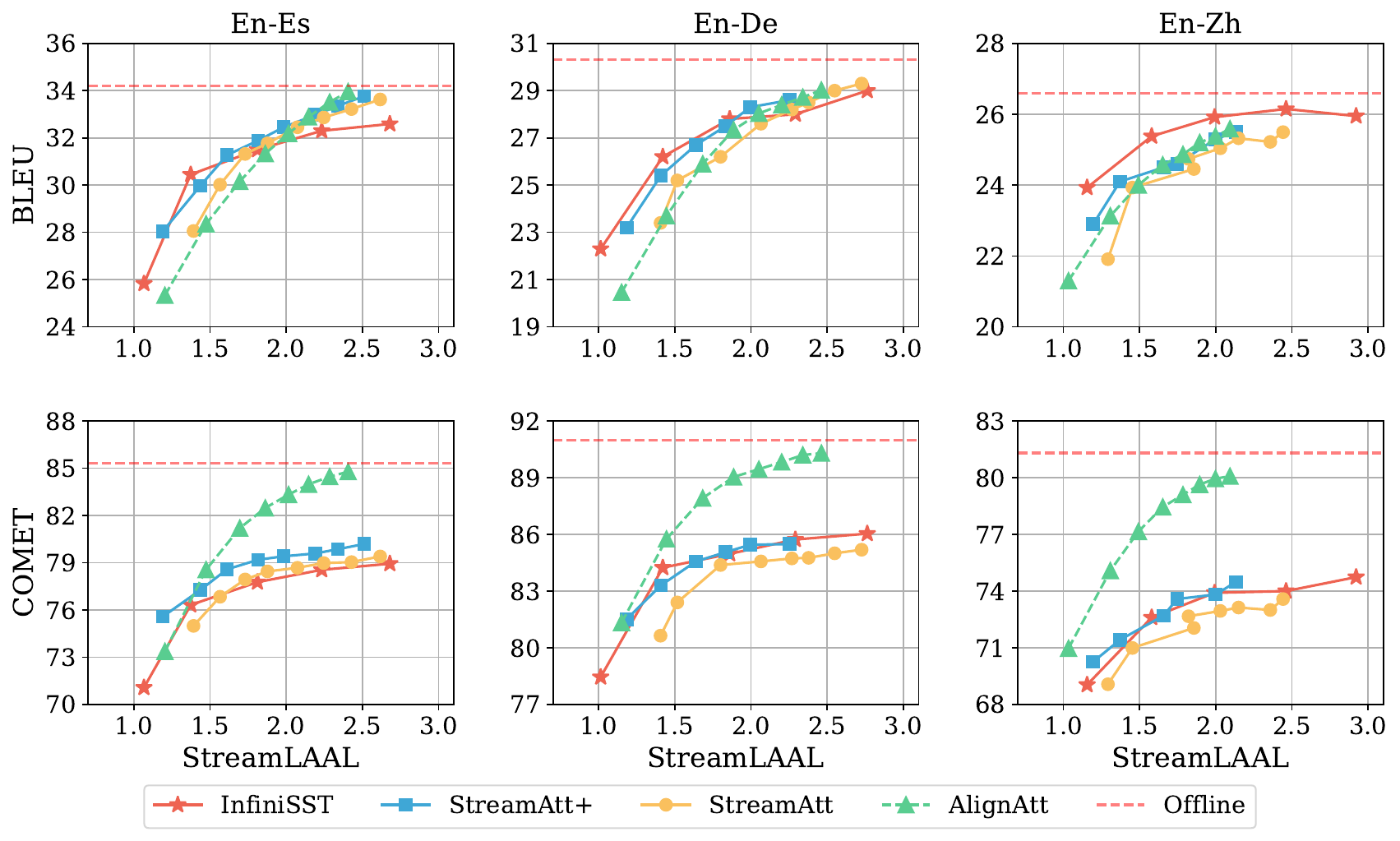}
    \caption{
    Quality-latency trade-off of \method~compared to the baselines on complete TED talks from the MuST-C tst-COMMON dataset in the En-Es, En-De, and En-Zh directions. Translation quality is measured using BLEU and COMET scores, while latency is evaluated using the \textit{non-computation-aware} StreamLAAL metric. For reference, we also include offline translation quality and results from AlignAtt tested on segmented speech. \method~achieves slightly better translation quality than StreamAtt at latency $\leq$ 1.5 seconds and remains competitive at higher latency levels.}
    \label{fig:main}
\end{figure*}

\begin{table}[t]
    \centering
    \begin{tabular}{c|c}\toprule
        Language & LAAL \\\midrule
        En-Zh & 1171 ms \\
        En-De & 924 ms \\
        En-Es & 850 ms\\\bottomrule
    \end{tabular}
    \caption{The latency of constructed trajectories for MuST-C En-Zh/De/Es sentence-level utterances evaluated with LAAL.}
    \label{tab:latency_traj}
\end{table}

\subsection{Data}

We conduct experiments on the En-Es (v1), En-De (v1), and En-Zh (v2) directions of the MuST-C dataset~\cite{di-gangi-etal-2019-must}. Due to the poor alignment quality in the En-Zh training set, we filter out misaligned ST triplets using CometKiWi~\cite{rei-etal-2022-cometkiwi} and retranslate them using TowerInstruct~\cite{tower}. 
Then, we construct trajectories and robust segments as described in Section~\ref{sec:data}. Examples of trajectory are shown in Figure \ref{fig:traj_eg}. The latency of trajectories are shown in Table \ref{tab:latency_traj}. Further statistics can be found in the Appendix \ref{sec:appdx_data}.

\subsection{Training} 

For all three language directions, we set maximum latency multiplier $M=12$. 
We adopt a two-stage supervised fine-tuning approach. 
In the first stage, we freeze the LLM and train only the speech encoder and adapter for 6 epochs with an effective batch size of 57.6K tokens. We use Adam optimizer~\cite{kingma2017adammethodstochasticoptimization} with learning rate \( 2 \times 10^{-4} \) and 1000 warmup steps.
In the second stage, we fine-tune the entire LLM for 1 epoch with an effective batch size of 76.8K tokens, a learning rate of \( 7 \times 10^{-6} \) and 100 warmup steps. 
We apply gradient clipping with a norm of 1.0.
We employ DeepSpeed Zero Stage-2 optimization\footnote{\url{https://github.com/deepspeedai/DeepSpeed}}, and enable optimizer and parameter offloading during the second training stage.
All models referred to in this paper are trained on a single node of 8 L40S GPU.

\subsection{Inference}

We use beam search with beam width 4 for all methods. 
We set \texttt{no\_repeat\_ngram\_size=5} and \texttt{repetition\_penalty=1.2} to suppress repetition\footnote{\url{https://huggingface.co/docs/transformers/en/main_classes/text_generation\#transformers.GenerationConfig}}. 
The sliding window size of LLM is set to 1000 tokens and that of speech encoder is set to 10 speech chunks. 
We vary the latency multiplier $m$ from 1 to 5 for \method~to obtain a quality-latency trade-off. 

\subsection{Evaluation}

We evaluate SST on complete TED Talks from the MuST-C tst-COMMON set, which consists of 27 TED Talks with durations ranging from 3 to 23 minutes.
To assess translation quality, we use SacreBLEU~\cite{post-2018-call} and COMET~\cite{guerreiro-etal-2024-xcomet}. Following the WMT24 practice~\cite{freitag-etal-2024-llms}, we compute the COMET score by averaging the scores from XCOMET-XL and XCOMET-XXL.
For latency evaluation, we use Length-Adaptive Average Lagging (LAAL)~\cite{papi-etal-2022-generation} for segmented speech baselines and StreamLAAL~\cite{papi-etal-2024-streamatt} for unbounded speech, both implemented within the SimulEval framework~\cite{ma-etal-2020-simuleval}. 
Computation cost is measured using both computation-aware StreamLAAL (StreamLAAL\_CA) and the Real-Time Factor (RTF), defined as the ratio of wall-clock computation time to speech duration.

\subsection{Baselines}

We compare our method against the following baselines:

\paragraph{AlignAtt}~\cite{alignatt} 
is a state-of-the-art SST policy applied to offline ST models, translating based on attention scores between translation outputs and speech utterances. It is designed for SST on segmented speech, and we include its results as a reference. We train an offline ST model using segmented ST triplets from MuST-C and robust segments that we constructed. 
The offline ST model uses the LST architecture~\cite{lst}, where the LLM inputs are organized as (instruction, speech history, translation history) rather than interleaving speech and translation as in~\method. We use the attention scores from layer 14 of the LLM and vary the number of frames from 1 to 8. 

\paragraph{StreamAtt}~\cite{papi-etal-2024-streamatt} 
extends AlignAtt to unbounded speech by maintaining both text and audio history through attention-based selection. We adopt the Fixed-Word approach from StreamAtt, preserving 40 words in the text history. To prevent excessively long preserved speech, we apply truncation when the duration exceeds 28.8 seconds.

\paragraph{StreamAtt+} We observe that the vanilla truncation strategy sometimes removes too much audio, leading to critical misalignment between the preserved speech and its translation. To mitigate this issue, we modify StreamAtt by ensuring that audio segments shorter than 10 seconds are never truncated.

\section{Main Results}

\paragraph{Lower Computation Cost} We run all inference experiments on a single NVIDIA L40S GPU and an AMD EPYC 9354 32-Core CPU. Results evaluated with StreamLAAL\_CA are shown in Figure \ref{fig:main_ca}. 
\method~achieves 0.5 to 1 second lower computation aware latency compared to StreamAtt and StreamAtt+ at the same quality level. We also compare the Real-Time Factor (RTF) of \method~and StreamAtt+ in Figure~\ref{fig:rtf}.  The RTF of \method~is significantly lower than StreamAtt+, indicating that the computation overhead of \method~is less than half of the StreamAtt+.

\paragraph{Competitive Translation Quality at the Same Theoretical Latency} Results evaluated with non-computation-aware StreamLAAL are shown in Figure~\ref{fig:main}. When StreamLAAL is no more than 1.5 second, \method~achieves slightly higher BLEU scores ($0.5\sim 1.0$) and similar COMET scores than StreamAtt+ on all three language directions. When StreamLAAL is more than 1.5 second, \method~still achieves higher BLEU score on En-Zh direction and competitive with StreamAtt+ on the En-De and En-Es directions. We note that AlignAtt tested on segmented speech exhibit significant higher COMET scores but not BLEU scores than both \method~and StreamAtt on all three language directions. A possible reason is that StreamLAAL uses mWERSegmenter~\cite{matusov-etal-2005-evaluating} to find alignment between translation of the complete talk and segmented references, and COMET is more sensitive to such misalignment than BLEU.

\section{Ablation Studies}

The default model we use in the ablation study is trained with robust segments and a maximum latency multiplier of \( M = 4 \) on the En-Zh direction.

\begin{table}[t]
    \centering
    \begin{tabular}{c|c|c}\toprule
        Robust  & Non-Robust  & Non-Robust   \\
        Segments & Segments & Segments\textsuperscript{*} \\ \midrule
        69.2 / 1.1 & 50.5 / -220 & 51.0 / -207 \\
        71.9 / 1.5 & 53.4 / -116 & 58.1 / -58\enspace \\
        72.3 / 1.9 & 68.4  / 2\enspace\enspace\; & 65.7 / -22\enspace \\
        73.0 / 2.4 & 66.8 / -12\enspace & 67.2 / -6\enspace\enspace \\\bottomrule
    \end{tabular}
    \caption{Impact of robust segments evaluated on MuST-C En-Zh tst-COMMON with latency multipliers $m=1,2,3,4$. A / B stands for COMET / LAAL (in second). The model trained on non-robust segments fails to translate unbounded speech. \textsuperscript{*}We suppress the non-linguistic sound tokens but still the model fails to generalize. }
    \label{tab:mega_chunk}
\end{table}

\subsection{Data}

\paragraph{Robust Segments}

We evaluate the effectiveness of robust segments by comparing \method~trained on trajectories of robust segments with \method~trained on trajectories of original MuST-C segmented speech. Both models are evaluated on tst-COMMON En-Zh with latency multipliers \( m \in [1,4] \), and the results are presented in Table~\ref{tab:mega_chunk}. 

The model trained on trajectories of non-robust segments exhibits abnormal latency scores and lower translation quality compared to the model trained on trajectories of robust segments. Manual examination of translation instances reveals that the segmented speech model frequently falls into repetition of non-linguistic tokens such as \begin{CJK*}{UTF8}{gbsn}（笑声）\end{CJK*} (meaning laughter) whenever non-linguistic sounds appear in the audio. 

We attempted to suppress these tokens, and the results are reported in the last column of Table~\ref{tab:mega_chunk}. Instead of producing repetitive tokens, the model stops generating translations upon encountering non-linguistic sounds. These findings highlight the importance of training with robust segments.

\begin{figure}[t]
    \centering
    \includegraphics[width=0.8\linewidth]{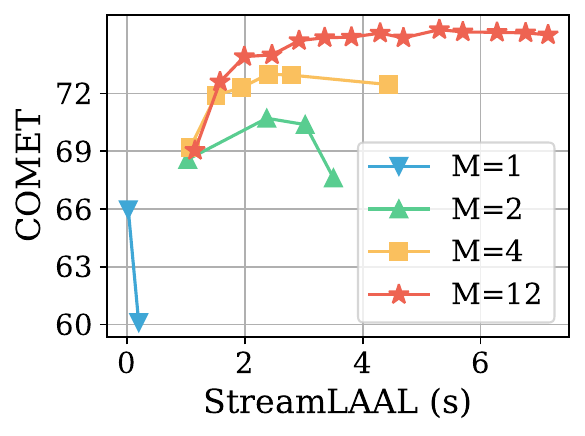}
    \caption{\method~trained with maximum latency multipliers $M=1,2,4,12$ and evaluated with $m\leq M + 2$. Larger maximum latency multipliers during training lead to improved quality-latency trade-offs.}
    \label{fig:ablation_data_mult}
\end{figure}

\paragraph{Multi-Latency}

We evaluate the effectiveness of multi-latency augmentation by training models with maximum latency multipliers $M = 1, 2, 4$, and $12$, and performing inference with $m \leq M + 2$. Results are shown in Figure~\ref{fig:ablation_data_mult}.

Larger $M$ consistently leads to better quality-latency trade-offs. Within the training range ($m \leq M$), translation quality improves with higher $m$; beyond it ($m > M$), quality degrades since it is out of the model's training distribution. However, models trained with larger $M$ degrade less when extrapolated to $m > M$.

These results highlight the importance of training with a sufficiently large $M$ while keeping $m \leq M$ at inference for optimal performance.

\subsection{Speech Encoder}

\paragraph{Inference Cache Window} 
We first evaluate how the speech encoder's cache window during inference affects model performance. The model is trained with \( w^s = 10 \) and tested with \( w^s = 5, 10, 20, \) and \( 40 \). The results, presented in Table~\ref{tab:infer_cache}, indicate that using a different cache window size during inference than the one used during training degrades translation quality.

\paragraph{Training Cache Window} Furthermore, we train models with different cache window sizes \( w^s = 10, 20, 30 \) while ensuring that the cache window size matches between training and inference. Since each robust segment has a size of 30, training with \( w^s = 30 \) disables the sliding window mechanism. The results, shown in Figure~\ref{fig:train_cache_speech}, reveal a surprising observation: the model trained with \( w^s = 30 \) successfully scales to unbounded speech during inference despite not using a sliding window during training. It also achieves a slightly better quality-latency trade-off compared to the model trained with \( w^s = 10 \).
These findings suggest using the largest possible speech cache window that GPU memory allows. 

\begin{table}[t]
    \centering
    \resizebox{\linewidth}{!}{
    \begin{tabular}{c|c|c}\toprule
        Speech Cache  & LLM Cache & \multirow{2}{*}{Quality / Latency} \\
        Window $w^s$ & Window $w^t$ & \\\midrule
        10 & 1000 & 69.2 / 1.1 \\\midrule
        5 & \multirow{3}{*}{1000} & 68.7 / 1.1 \\
        20 & & 68.3 / 1.0 \\
        40 & & 66.1 / 0.9 \\\midrule
        \multirow{3}{*}{10} & 500 & 69.0 / 1.0 \\
        & 2000 & 69.4 / 1.2 \\
        & 4000 & 69.4 / 1.2 \\\bottomrule
    \end{tabular}
    }    
    \caption{Impact of cache size during inference. Quality is evaluated with COMET and latency is evaluated with StreamLAAL (unit is second). Model is trained with speech encoder sliding window $w^s=10$ and no sliding window for LLM. Latency multiplier is set to $m=1$.}
    \label{tab:infer_cache}
\end{table}

\subsection{LLM}

\paragraph{Cache Instruction} 
As described in Section~\ref{sec:infer}, we explicitly preserve the KV cache of the translation instruction at the beginning (i.e., the system prompt). If this cache is not retained, the LLM stops translating once the window starts sliding. 

\paragraph{Cache Window \( w^t \)} 
We evaluate the impact of the LLM's cache window size during inference on model performance. Notably, the sliding window mechanism is not applied to the LLM during training. We vary the LLM cache window size as \( w^t = 500, 1000, 2000, 4000 \), and the results are presented in Table~\ref{tab:infer_cache}. 
Increasing the KV cache size slightly improves translation quality (\( 69 \to 69.4 \)) at the cost of marginally higher latency (\( 1.0 \to 1.2 \)). Compared to the speech encoder, the LLM demonstrates greater robustness to different KV cache window sizes.


\begin{table}[t]
    \centering
    \resizebox{\linewidth}{!}{
    \begin{tabular}{c|c|c}\toprule
        Model & Talks $\leq 10$min & Talks $>10$min \\ \midrule
        Llama-3-8K & 70.9 / 1.0  & 67.1  / 1.1 \\
        Llama-3.1-128K & 71.6 / 1.0 & 68.0 / 1.1 \\\bottomrule
    \end{tabular}
    }
    \caption{Impact of LLM context length. A / B stands for COMET / LAAL (in second). Llama-3 with 8K context length is still able to generalize to talks longer than 10 minutes.}
    \label{tab:llm_context_length}
\end{table}

\paragraph{Base LLM Context Length} 
Throughout our experiments, we use Llama-3.1-8B-Instruct as the base LLM, which supports a context length of up to 128K tokens. To assess whether \method~generalizes to an LLM with a shorter context limit, we replace it with Llama-3-8B-Instruct, which has an 8K context length\footnote{A 10-minute speech already generates \( 10 \cdot 60 \cdot 12.5 = 7.5K \) speech embeddings, exceeding the 8K context limit of Llama-3-8B-Instruct if combined with the translation tokens.}. 
The results, presented in Table~\ref{tab:llm_context_length}, indicate that while Llama-3 exhibits lower translation quality compared to Llama-3.1, it is still capable of generalizing to unbounded speech with \method.

\section{Conclusion}

We propose \method~that enables simultaneous translation of unbounded speech with state-of-the-art quality latency trade-off on three language directions of MuST-C dataset. Our ablations demonstrate the effectiveness of our carefully constructed data, including robust segments and multi-latency augmentation, and cache management strategy during inference. 

\section*{Limitations}

On the higher theoretical latency level, \method~still falls behind AlignAtt and StreamAtt in some cases. This can be attributed to the limited bidirectional attention of the chunkwise-causal speech encoder. Also, we evaluated on En-X directions but not on other directions like X-En and X-X. We have not experimented with other pretrained speech encoders and non-Llama LLMs due to computation budget. Besides, the StreamLAAL metric is not perfectly reliable due to alignment errors of mWERSegmenter. Finally, we have not conducted human evaluation on user experience of different SST models, which might reveal undetected flaws in current models. 

\bibliography{clean}

\begin{thebibliography}{43}
\providecommand{\natexlab}[1]{#1}

\bibitem[{Alves et~al.(2024)Alves, Pombal, Guerreiro, Martins, Alves, Farajian,
  Peters, Rei, Fernandes, Agrawal, Colombo, de~Souza, and Martins}]{tower}
Duarte~Miguel Alves, Jos{\'e} Pombal, Nuno~M Guerreiro, Pedro~Henrique Martins,
  Jo{\~a}o Alves, Amin Farajian, Ben Peters, Ricardo Rei, Patrick Fernandes,
  Sweta Agrawal, Pierre Colombo, Jos{\'e} G.~C. de~Souza, and Andre Martins.
  2024.
\newblock \href {https://openreview.net/forum?id=EHPns3hVkj} {Tower: An open
  multilingual large language model for translation-related tasks}.
\newblock In \emph{First Conference on Language Modeling}.

\bibitem[{Baevski et~al.(2020)Baevski, Zhou, Mohamed, and Auli}]{wav2vec2}
Alexei Baevski, Yuhao Zhou, Abdelrahman Mohamed, and Michael Auli. 2020.
\newblock \href {https://proceedings.neurips.cc/paper_files/paper/2020/file/
  92d1e1eb1cd6f9fba3227870bb6d7f07-Paper.pdf} {wav2vec 2.0: A framework for
  self-supervised learning of speech representations}.
\newblock In \emph{Advances in Neural Information Processing Systems},
  volume~33, pages 12449--12460. Curran Associates, Inc.

\bibitem[{Deng et~al.(2022)Deng, Watanabe, Shi, and Arora}]{blockcausal}
Keqi Deng, Shinji Watanabe, Jiatong Shi, and Siddhant Arora. 2022.
\newblock \href {https://doi.org/10.21437/Interspeech.2022-933} {Blockwise
  streaming transformer for spoken language understanding and simultaneous
  speech translation}.
\newblock In \emph{Interspeech 2022}, pages 1746--1750.

\bibitem[{Di~Gangi et~al.(2019)Di~Gangi, Cattoni, Bentivogli, Negri, and
  Turchi}]{di-gangi-etal-2019-must}
Mattia~A. Di~Gangi, Roldano Cattoni, Luisa Bentivogli, Matteo Negri, and Marco
  Turchi. 2019.
\newblock \href {https://doi.org/10.18653/v1/N19-1202} {{M}u{ST}-{C}: a
  {M}ultilingual {S}peech {T}ranslation {C}orpus}.
\newblock In \emph{Proceedings of the 2019 Conference of the North {A}merican
  Chapter of the Association for Computational Linguistics: Human Language
  Technologies, Volume 1 (Long and Short Papers)}, pages 2012--2017,
  Minneapolis, Minnesota. Association for Computational Linguistics.

\bibitem[{Donato et~al.(2021)Donato, Yu, and Dyer}]{donato-etal-2021-diverse}
Domenic Donato, Lei Yu, and Chris Dyer. 2021.
\newblock \href {https://doi.org/10.18653/v1/2021.acl-long.104} {Diverse
  pretrained context encodings improve document translation}.
\newblock In \emph{Proceedings of the 59th Annual Meeting of the Association
  for Computational Linguistics and the 11th International Joint Conference on
  Natural Language Processing (Volume 1: Long Papers)}, pages 1299--1311,
  Online. Association for Computational Linguistics.

\bibitem[{Dong et~al.(2022)Dong, Zhu, Wang, and Li}]{dong-etal-2022-learning}
Qian Dong, Yaoming Zhu, Mingxuan Wang, and Lei Li. 2022.
\newblock \href {https://doi.org/10.18653/v1/2022.acl-long.50} {Learning when
  to translate for streaming speech}.
\newblock In \emph{Proceedings of the 60th Annual Meeting of the Association
  for Computational Linguistics (Volume 1: Long Papers)}, pages 680--694,
  Dublin, Ireland. Association for Computational Linguistics.

\bibitem[{Feng et~al.(2022)Feng, Yang, Cer, Arivazhagan, and
  Wang}]{feng-etal-2022-language}
Fangxiaoyu Feng, Yinfei Yang, Daniel Cer, Naveen Arivazhagan, and Wei Wang.
  2022.
\newblock \href {https://doi.org/10.18653/v1/2022.acl-long.62}
  {Language-agnostic {BERT} sentence embedding}.
\newblock In \emph{Proceedings of the 60th Annual Meeting of the Association
  for Computational Linguistics (Volume 1: Long Papers)}, pages 878--891,
  Dublin, Ireland. Association for Computational Linguistics.

\bibitem[{Freitag et~al.(2024)Freitag, Mathur, Deutsch, Lo, Avramidis, Rei,
  Thompson, Blain, Kocmi, Wang, Adelani, Buchicchio, Zerva, and
  Lavie}]{freitag-etal-2024-llms}
Markus Freitag, Nitika Mathur, Daniel Deutsch, Chi-Kiu Lo, Eleftherios
  Avramidis, Ricardo Rei, Brian Thompson, Frederic Blain, Tom Kocmi, Jiayi
  Wang, David~Ifeoluwa Adelani, Marianna Buchicchio, Chrysoula Zerva, and Alon
  Lavie. 2024.
\newblock \href {https://doi.org/10.18653/v1/2024.wmt-1.2} {Are {LLM}s breaking
  {MT} metrics? results of the {WMT}24 metrics shared task}.
\newblock In \emph{Proceedings of the Ninth Conference on Machine Translation},
  pages 47--81, Miami, Florida, USA. Association for Computational Linguistics.

\bibitem[{Fugen et~al.(2006)Fugen, Kolss, Bernreuther, Paulik, Stuker, Vogel,
  and Waibel}]{1660084}
C.~Fugen, M.~Kolss, D.~Bernreuther, M.~Paulik, S.~Stuker, S.~Vogel, and
  A.~Waibel. 2006.
\newblock \href {https://doi.org/10.1109/ICASSP.2006.1660084} {Open domain
  speech recognition \& translation:lectures and speeches}.
\newblock In \emph{2006 IEEE International Conference on Acoustics Speech and
  Signal Processing Proceedings}, volume~1, pages I--I.

\bibitem[{Grattafiori et~al.(2024)Grattafiori, Dubey, Jauhri, Pandey, Kadian,
  Al-Dahle, Letman, Mathur, Schelten, Vaughan, Yang, Fan, Goyal, Hartshorn,
  Yang, Mitra, Sravankumar, Korenev, Hinsvark, Rao, Zhang, Rodriguez,
  Gregerson, Spataru, Roziere, Biron, Tang, Chern, Caucheteux, Nayak, Bi,
  Marra, McConnell, Keller, Touret, Wu, Wong, Ferrer, Nikolaidis, Allonsius,
  Song, Pintz, Livshits, Wyatt, Esiobu, Choudhary, Mahajan, Garcia-Olano,
  Perino, Hupkes, Lakomkin, AlBadawy, Lobanova, Dinan, Smith, Radenovic,
  Guzmán, Zhang, Synnaeve, Lee, Anderson, Thattai, Nail, Mialon, Pang,
  Cucurell, Nguyen, Korevaar, Xu, Touvron, Zarov, Ibarra, Kloumann, Misra,
  Evtimov, Zhang, Copet, Lee, Geffert, Vranes, Park, Mahadeokar, Shah, van~der
  Linde, Billock, Hong, Lee, Fu, Chi, Huang, Liu, Wang, Yu, Bitton, Spisak,
  Park, Rocca, Johnstun, Saxe, Jia, Alwala, Prasad, Upasani, Plawiak, Li,
  Heafield, Stone, El-Arini, Iyer, Malik, Chiu, Bhalla, Lakhotia,
  Rantala-Yeary, van~der Maaten, Chen, Tan, Jenkins, Martin, Madaan, Malo,
  Blecher, Landzaat, de~Oliveira, Muzzi, Pasupuleti, Singh, Paluri, Kardas,
  Tsimpoukelli, Oldham, Rita, Pavlova, Kambadur, Lewis, Si, Singh, Hassan,
  Goyal, Torabi, Bashlykov, Bogoychev, Chatterji, Zhang, Duchenne, Çelebi,
  Alrassy, Zhang, Li, Vasic, Weng, Bhargava, Dubal, Krishnan, Koura, Xu, He,
  Dong, Srinivasan, Ganapathy, Calderer, Cabral, Stojnic, Raileanu, Maheswari,
  Girdhar, Patel, Sauvestre, Polidoro, Sumbaly, Taylor, Silva, Hou, Wang,
  Hosseini, Chennabasappa, Singh, Bell, Kim, Edunov, Nie, Narang, Raparthy,
  Shen, Wan, Bhosale, Zhang, Vandenhende, Batra, Whitman, Sootla, Collot,
  Gururangan, Borodinsky, Herman, Fowler, Sheasha, Georgiou, Scialom,
  Speckbacher, Mihaylov, Xiao, Karn, Goswami, Gupta, Ramanathan, Kerkez,
  Gonguet, Do, Vogeti, Albiero, Petrovic, Chu, Xiong, Fu, Meers, Martinet,
  Wang, Wang, Tan, Xia, Xie, Jia, Wang, Goldschlag, Gaur, Babaei, Wen, Song,
  Zhang, Li, Mao, Coudert, Yan, Chen, Papakipos, Singh, Srivastava, Jain,
  Kelsey, Shajnfeld, Gangidi, Victoria, Goldstand, Menon, Sharma, Boesenberg,
  Baevski, Feinstein, Kallet, Sangani, Teo, Yunus, Lupu, Alvarado, Caples, Gu,
  Ho, Poulton, Ryan, Ramchandani, Dong, Franco, Goyal, Saraf, Chowdhury,
  Gabriel, Bharambe, Eisenman, Yazdan, James, Maurer, Leonhardi, Huang, Loyd,
  Paola, Paranjape, Liu, Wu, Ni, Hancock, Wasti, Spence, Stojkovic, Gamido,
  Montalvo, Parker, Burton, Mejia, Liu, Wang, Kim, Zhou, Hu, Chu, Cai, Tindal,
  Feichtenhofer, Gao, Civin, Beaty, Kreymer, Li, Adkins, Xu, Testuggine, David,
  Parikh, Liskovich, Foss, Wang, Le, Holland, Dowling, Jamil, Montgomery,
  Presani, Hahn, Wood, Le, Brinkman, Arcaute, Dunbar, Smothers, Sun, Kreuk,
  Tian, Kokkinos, Ozgenel, Caggioni, Kanayet, Seide, Florez, Schwarz, Badeer,
  Swee, Halpern, Herman, Sizov, Guangyi, Zhang, Lakshminarayanan, Inan,
  Shojanazeri, Zou, Wang, Zha, Habeeb, Rudolph, Suk, Aspegren, Goldman, Zhan,
  Damlaj, Molybog, Tufanov, Leontiadis, Veliche, Gat, Weissman, Geboski, Kohli,
  Lam, Asher, Gaya, Marcus, Tang, Chan, Zhen, Reizenstein, Teboul, Zhong, Jin,
  Yang, Cummings, Carvill, Shepard, McPhie, Torres, Ginsburg, Wang, Wu, U,
  Saxena, Khandelwal, Zand, Matosich, Veeraraghavan, Michelena, Li, Jagadeesh,
  Huang, Chawla, Huang, Chen, Garg, A, Silva, Bell, Zhang, Guo, Yu, Moshkovich,
  Wehrstedt, Khabsa, Avalani, Bhatt, Mankus, Hasson, Lennie, Reso, Groshev,
  Naumov, Lathi, Keneally, Liu, Seltzer, Valko, Restrepo, Patel, Vyatskov,
  Samvelyan, Clark, Macey, Wang, Hermoso, Metanat, Rastegari, Bansal,
  Santhanam, Parks, White, Bawa, Singhal, Egebo, Usunier, Mehta, Laptev, Dong,
  Cheng, Chernoguz, Hart, Salpekar, Kalinli, Kent, Parekh, Saab, Balaji,
  Rittner, Bontrager, Roux, Dollar, Zvyagina, Ratanchandani, Yuvraj, Liang,
  Alao, Rodriguez, Ayub, Murthy, Nayani, Mitra, Parthasarathy, Li, Hogan,
  Battey, Wang, Howes, Rinott, Mehta, Siby, Bondu, Datta, Chugh, Hunt, Dhillon,
  Sidorov, Pan, Mahajan, Verma, Yamamoto, Ramaswamy, Lindsay, Lindsay, Feng,
  Lin, Zha, Patil, Shankar, Zhang, Zhang, Wang, Agarwal, Sajuyigbe, Chintala,
  Max, Chen, Kehoe, Satterfield, Govindaprasad, Gupta, Deng, Cho, Virk,
  Subramanian, Choudhury, Goldman, Remez, Glaser, Best, Koehler, Robinson, Li,
  Zhang, Matthews, Chou, Shaked, Vontimitta, Ajayi, Montanez, Mohan, Kumar,
  Mangla, Ionescu, Poenaru, Mihailescu, Ivanov, Li, Wang, Jiang, Bouaziz,
  Constable, Tang, Wu, Wang, Wu, Gao, Kleinman, Chen, Hu, Jia, Qi, Li, Zhang,
  Zhang, Adi, Nam, Yu, Wang, Zhao, Hao, Qian, Li, He, Rait, DeVito, Rosnbrick,
  Wen, Yang, Zhao, and Ma}]{llama3}
Aaron Grattafiori, Abhimanyu Dubey, Abhinav Jauhri, Abhinav Pandey, Abhishek
  Kadian, Ahmad Al-Dahle, Aiesha Letman, Akhil Mathur, Alan Schelten, Alex
  Vaughan, Amy Yang, Angela Fan, Anirudh Goyal, Anthony Hartshorn, Aobo Yang,
  Archi Mitra, Archie Sravankumar, Artem Korenev, Arthur Hinsvark, Arun Rao,
  Aston Zhang, Aurelien Rodriguez, Austen Gregerson, Ava Spataru, Baptiste
  Roziere, Bethany Biron, Binh Tang, Bobbie Chern, Charlotte Caucheteux, Chaya
  Nayak, Chloe Bi, Chris Marra, Chris McConnell, Christian Keller, Christophe
  Touret, Chunyang Wu, Corinne Wong, Cristian~Canton Ferrer, Cyrus Nikolaidis,
  Damien Allonsius, Daniel Song, Danielle Pintz, Danny Livshits, Danny Wyatt,
  David Esiobu, Dhruv Choudhary, Dhruv Mahajan, Diego Garcia-Olano, Diego
  Perino, Dieuwke Hupkes, Egor Lakomkin, Ehab AlBadawy, Elina Lobanova, Emily
  Dinan, Eric~Michael Smith, Filip Radenovic, Francisco Guzmán, Frank Zhang,
  Gabriel Synnaeve, Gabrielle Lee, Georgia~Lewis Anderson, Govind Thattai,
  Graeme Nail, Gregoire Mialon, Guan Pang, Guillem Cucurell, Hailey Nguyen,
  Hannah Korevaar, Hu~Xu, Hugo Touvron, Iliyan Zarov, Imanol~Arrieta Ibarra,
  Isabel Kloumann, Ishan Misra, Ivan Evtimov, Jack Zhang, Jade Copet, Jaewon
  Lee, Jan Geffert, Jana Vranes, Jason Park, Jay Mahadeokar, Jeet Shah, Jelmer
  van~der Linde, Jennifer Billock, Jenny Hong, Jenya Lee, Jeremy Fu, Jianfeng
  Chi, Jianyu Huang, Jiawen Liu, Jie Wang, Jiecao Yu, Joanna Bitton, Joe
  Spisak, Jongsoo Park, Joseph Rocca, Joshua Johnstun, Joshua Saxe, Junteng
  Jia, Kalyan~Vasuden Alwala, Karthik Prasad, Kartikeya Upasani, Kate Plawiak,
  Ke~Li, Kenneth Heafield, Kevin Stone, Khalid El-Arini, Krithika Iyer, Kshitiz
  Malik, Kuenley Chiu, Kunal Bhalla, Kushal Lakhotia, Lauren Rantala-Yeary,
  Laurens van~der Maaten, Lawrence Chen, Liang Tan, Liz Jenkins, Louis Martin,
  Lovish Madaan, Lubo Malo, Lukas Blecher, Lukas Landzaat, Luke de~Oliveira,
  Madeline Muzzi, Mahesh Pasupuleti, Mannat Singh, Manohar Paluri, Marcin
  Kardas, Maria Tsimpoukelli, Mathew Oldham, Mathieu Rita, Maya Pavlova,
  Melanie Kambadur, Mike Lewis, Min Si, Mitesh~Kumar Singh, Mona Hassan, Naman
  Goyal, Narjes Torabi, Nikolay Bashlykov, Nikolay Bogoychev, Niladri
  Chatterji, Ning Zhang, Olivier Duchenne, Onur Çelebi, Patrick Alrassy,
  Pengchuan Zhang, Pengwei Li, Petar Vasic, Peter Weng, Prajjwal Bhargava,
  Pratik Dubal, Praveen Krishnan, Punit~Singh Koura, Puxin Xu, Qing He,
  Qingxiao Dong, Ragavan Srinivasan, Raj Ganapathy, Ramon Calderer,
  Ricardo~Silveira Cabral, Robert Stojnic, Roberta Raileanu, Rohan Maheswari,
  Rohit Girdhar, Rohit Patel, Romain Sauvestre, Ronnie Polidoro, Roshan
  Sumbaly, Ross Taylor, Ruan Silva, Rui Hou, Rui Wang, Saghar Hosseini, Sahana
  Chennabasappa, Sanjay Singh, Sean Bell, Seohyun~Sonia Kim, Sergey Edunov,
  Shaoliang Nie, Sharan Narang, Sharath Raparthy, Sheng Shen, Shengye Wan,
  Shruti Bhosale, Shun Zhang, Simon Vandenhende, Soumya Batra, Spencer Whitman,
  Sten Sootla, Stephane Collot, Suchin Gururangan, Sydney Borodinsky, Tamar
  Herman, Tara Fowler, Tarek Sheasha, Thomas Georgiou, Thomas Scialom, Tobias
  Speckbacher, Todor Mihaylov, Tong Xiao, Ujjwal Karn, Vedanuj Goswami, Vibhor
  Gupta, Vignesh Ramanathan, Viktor Kerkez, Vincent Gonguet, Virginie Do, Vish
  Vogeti, Vítor Albiero, Vladan Petrovic, Weiwei Chu, Wenhan Xiong, Wenyin Fu,
  Whitney Meers, Xavier Martinet, Xiaodong Wang, Xiaofang Wang, Xiaoqing~Ellen
  Tan, Xide Xia, Xinfeng Xie, Xuchao Jia, Xuewei Wang, Yaelle Goldschlag,
  Yashesh Gaur, Yasmine Babaei, Yi~Wen, Yiwen Song, Yuchen Zhang, Yue Li,
  Yuning Mao, Zacharie~Delpierre Coudert, Zheng Yan, Zhengxing Chen, Zoe
  Papakipos, Aaditya Singh, Aayushi Srivastava, Abha Jain, Adam Kelsey, Adam
  Shajnfeld, Adithya Gangidi, Adolfo Victoria, Ahuva Goldstand, Ajay Menon,
  Ajay Sharma, Alex Boesenberg, Alexei Baevski, Allie Feinstein, Amanda Kallet,
  Amit Sangani, Amos Teo, Anam Yunus, Andrei Lupu, Andres Alvarado, Andrew
  Caples, Andrew Gu, Andrew Ho, Andrew Poulton, Andrew Ryan, Ankit Ramchandani,
  Annie Dong, Annie Franco, Anuj Goyal, Aparajita Saraf, Arkabandhu Chowdhury,
  Ashley Gabriel, Ashwin Bharambe, Assaf Eisenman, Azadeh Yazdan, Beau James,
  Ben Maurer, Benjamin Leonhardi, Bernie Huang, Beth Loyd, Beto~De Paola,
  Bhargavi Paranjape, Bing Liu, Bo~Wu, Boyu Ni, Braden Hancock, Bram Wasti,
  Brandon Spence, Brani Stojkovic, Brian Gamido, Britt Montalvo, Carl Parker,
  Carly Burton, Catalina Mejia, Ce~Liu, Changhan Wang, Changkyu Kim, Chao Zhou,
  Chester Hu, Ching-Hsiang Chu, Chris Cai, Chris Tindal, Christoph
  Feichtenhofer, Cynthia Gao, Damon Civin, Dana Beaty, Daniel Kreymer, Daniel
  Li, David Adkins, David Xu, Davide Testuggine, Delia David, Devi Parikh,
  Diana Liskovich, Didem Foss, Dingkang Wang, Duc Le, Dustin Holland, Edward
  Dowling, Eissa Jamil, Elaine Montgomery, Eleonora Presani, Emily Hahn, Emily
  Wood, Eric-Tuan Le, Erik Brinkman, Esteban Arcaute, Evan Dunbar, Evan
  Smothers, Fei Sun, Felix Kreuk, Feng Tian, Filippos Kokkinos, Firat Ozgenel,
  Francesco Caggioni, Frank Kanayet, Frank Seide, Gabriela~Medina Florez,
  Gabriella Schwarz, Gada Badeer, Georgia Swee, Gil Halpern, Grant Herman,
  Grigory Sizov, Guangyi, Zhang, Guna Lakshminarayanan, Hakan Inan, Hamid
  Shojanazeri, Han Zou, Hannah Wang, Hanwen Zha, Haroun Habeeb, Harrison
  Rudolph, Helen Suk, Henry Aspegren, Hunter Goldman, Hongyuan Zhan, Ibrahim
  Damlaj, Igor Molybog, Igor Tufanov, Ilias Leontiadis, Irina-Elena Veliche,
  Itai Gat, Jake Weissman, James Geboski, James Kohli, Janice Lam, Japhet
  Asher, Jean-Baptiste Gaya, Jeff Marcus, Jeff Tang, Jennifer Chan, Jenny Zhen,
  Jeremy Reizenstein, Jeremy Teboul, Jessica Zhong, Jian Jin, Jingyi Yang, Joe
  Cummings, Jon Carvill, Jon Shepard, Jonathan McPhie, Jonathan Torres, Josh
  Ginsburg, Junjie Wang, Kai Wu, Kam~Hou U, Karan Saxena, Kartikay Khandelwal,
  Katayoun Zand, Kathy Matosich, Kaushik Veeraraghavan, Kelly Michelena, Keqian
  Li, Kiran Jagadeesh, Kun Huang, Kunal Chawla, Kyle Huang, Lailin Chen,
  Lakshya Garg, Lavender A, Leandro Silva, Lee Bell, Lei Zhang, Liangpeng Guo,
  Licheng Yu, Liron Moshkovich, Luca Wehrstedt, Madian Khabsa, Manav Avalani,
  Manish Bhatt, Martynas Mankus, Matan Hasson, Matthew Lennie, Matthias Reso,
  Maxim Groshev, Maxim Naumov, Maya Lathi, Meghan Keneally, Miao Liu,
  Michael~L. Seltzer, Michal Valko, Michelle Restrepo, Mihir Patel, Mik
  Vyatskov, Mikayel Samvelyan, Mike Clark, Mike Macey, Mike Wang, Miquel~Jubert
  Hermoso, Mo~Metanat, Mohammad Rastegari, Munish Bansal, Nandhini Santhanam,
  Natascha Parks, Natasha White, Navyata Bawa, Nayan Singhal, Nick Egebo,
  Nicolas Usunier, Nikhil Mehta, Nikolay~Pavlovich Laptev, Ning Dong, Norman
  Cheng, Oleg Chernoguz, Olivia Hart, Omkar Salpekar, Ozlem Kalinli, Parkin
  Kent, Parth Parekh, Paul Saab, Pavan Balaji, Pedro Rittner, Philip Bontrager,
  Pierre Roux, Piotr Dollar, Polina Zvyagina, Prashant Ratanchandani, Pritish
  Yuvraj, Qian Liang, Rachad Alao, Rachel Rodriguez, Rafi Ayub, Raghotham
  Murthy, Raghu Nayani, Rahul Mitra, Rangaprabhu Parthasarathy, Raymond Li,
  Rebekkah Hogan, Robin Battey, Rocky Wang, Russ Howes, Ruty Rinott, Sachin
  Mehta, Sachin Siby, Sai~Jayesh Bondu, Samyak Datta, Sara Chugh, Sara Hunt,
  Sargun Dhillon, Sasha Sidorov, Satadru Pan, Saurabh Mahajan, Saurabh Verma,
  Seiji Yamamoto, Sharadh Ramaswamy, Shaun Lindsay, Shaun Lindsay, Sheng Feng,
  Shenghao Lin, Shengxin~Cindy Zha, Shishir Patil, Shiva Shankar, Shuqiang
  Zhang, Shuqiang Zhang, Sinong Wang, Sneha Agarwal, Soji Sajuyigbe, Soumith
  Chintala, Stephanie Max, Stephen Chen, Steve Kehoe, Steve Satterfield,
  Sudarshan Govindaprasad, Sumit Gupta, Summer Deng, Sungmin Cho, Sunny Virk,
  Suraj Subramanian, Sy~Choudhury, Sydney Goldman, Tal Remez, Tamar Glaser,
  Tamara Best, Thilo Koehler, Thomas Robinson, Tianhe Li, Tianjun Zhang, Tim
  Matthews, Timothy Chou, Tzook Shaked, Varun Vontimitta, Victoria Ajayi,
  Victoria Montanez, Vijai Mohan, Vinay~Satish Kumar, Vishal Mangla, Vlad
  Ionescu, Vlad Poenaru, Vlad~Tiberiu Mihailescu, Vladimir Ivanov, Wei Li,
  Wenchen Wang, Wenwen Jiang, Wes Bouaziz, Will Constable, Xiaocheng Tang,
  Xiaojian Wu, Xiaolan Wang, Xilun Wu, Xinbo Gao, Yaniv Kleinman, Yanjun Chen,
  Ye~Hu, Ye~Jia, Ye~Qi, Yenda Li, Yilin Zhang, Ying Zhang, Yossi Adi, Youngjin
  Nam, Yu, Wang, Yu~Zhao, Yuchen Hao, Yundi Qian, Yunlu Li, Yuzi He, Zach Rait,
  Zachary DeVito, Zef Rosnbrick, Zhaoduo Wen, Zhenyu Yang, Zhiwei Zhao, and
  Zhiyu Ma. 2024.
\newblock \href {https://arxiv.org/abs/2407.21783} {The llama 3 herd of
  models}.

\bibitem[{Guerreiro et~al.(2024)Guerreiro, Rei, Stigt, Coheur, Colombo, and
  Martins}]{guerreiro-etal-2024-xcomet}
Nuno~M. Guerreiro, Ricardo Rei, Daan~van Stigt, Luisa Coheur, Pierre Colombo,
  and Andr{\'e} F.~T. Martins. 2024.
\newblock \href {https://doi.org/10.1162/tacl_a_00683} {xcomet: Transparent
  machine translation evaluation through fine-grained error detection}.
\newblock \emph{Transactions of the Association for Computational Linguistics},
  12:979--995.

\bibitem[{Han et~al.(2024)Han, Wang, Peng, Xiong, Chen, Ji, and
  Wang}]{han-etal-2024-lm}
Chi Han, Qifan Wang, Hao Peng, Wenhan Xiong, Yu~Chen, Heng Ji, and Sinong Wang.
  2024.
\newblock \href {https://doi.org/10.18653/v1/2024.naacl-long.222}
  {{LM}-infinite: Zero-shot extreme length generalization for large language
  models}.
\newblock In \emph{Proceedings of the 2024 Conference of the North American
  Chapter of the Association for Computational Linguistics: Human Language
  Technologies (Volume 1: Long Papers)}, pages 3991--4008, Mexico City, Mexico.
  Association for Computational Linguistics.

\bibitem[{Huang et~al.(2022)Huang, yiin Chang, Rybach, Prabhavalkar, Sainath,
  Allauzen, Peyser, and Lu}]{huang2022e2esegmenterjointsegmenting}
W.~Ronny Huang, Shuo yiin Chang, David Rybach, Rohit Prabhavalkar, Tara~N.
  Sainath, Cyril Allauzen, Cal Peyser, and Zhiyun Lu. 2022.
\newblock \href {https://arxiv.org/abs/2204.10749} {E2e segmenter: Joint
  segmenting and decoding for long-form asr}.

\bibitem[{Iranzo-S{\'a}nchez et~al.(2024)Iranzo-S{\'a}nchez,
  Iranzo-S{\'a}nchez, Gim{\'e}nez, Civera, and
  Juan}]{iranzo-sanchez-etal-2024-segmentation}
Javier Iranzo-S{\'a}nchez, Jorge Iranzo-S{\'a}nchez, Adri{\`a} Gim{\'e}nez,
  Jorge Civera, and Alfons Juan. 2024.
\newblock \href {https://doi.org/10.1162/tacl_a_00691} {Segmentation-free
  streaming machine translation}.
\newblock \emph{Transactions of the Association for Computational Linguistics},
  12:1104--1121.

\bibitem[{Jalili~Sabet et~al.(2020)Jalili~Sabet, Dufter, Yvon, and
  Sch{\"u}tze}]{jalili-sabet-etal-2020-simalign}
Masoud Jalili~Sabet, Philipp Dufter, Fran{\c{c}}ois Yvon, and Hinrich
  Sch{\"u}tze. 2020.
\newblock \href {https://doi.org/10.18653/v1/2020.findings-emnlp.147}
  {{S}im{A}lign: High quality word alignments without parallel training data
  using static and contextualized embeddings}.
\newblock In \emph{Findings of the Association for Computational Linguistics:
  EMNLP 2020}, pages 1627--1643, Online. Association for Computational
  Linguistics.

\bibitem[{Kingma and Ba(2017)}]{kingma2017adammethodstochasticoptimization}
Diederik~P. Kingma and Jimmy Ba. 2017.
\newblock \href {https://arxiv.org/abs/1412.6980} {Adam: A method for
  stochastic optimization}.

\bibitem[{Liu et~al.(2021)Liu, Du, Li, Li, and Chen}]{liu-etal-2021-cross}
Dan Liu, Mengge Du, Xiaoxi Li, Ya~Li, and Enhong Chen. 2021.
\newblock \href {https://doi.org/10.18653/v1/2021.emnlp-main.4} {Cross
  attention augmented transducer networks for simultaneous translation}.
\newblock In \emph{Proceedings of the 2021 Conference on Empirical Methods in
  Natural Language Processing}, pages 39--55, Online and Punta Cana, Dominican
  Republic. Association for Computational Linguistics.

\bibitem[{Ma et~al.(2020{\natexlab{a}})Ma, Dousti, Wang, Gu, and
  Pino}]{ma-etal-2020-simuleval}
Xutai Ma, Mohammad~Javad Dousti, Changhan Wang, Jiatao Gu, and Juan Pino.
  2020{\natexlab{a}}.
\newblock \href {https://doi.org/10.18653/v1/2020.emnlp-demos.19} {{SIMULEVAL}:
  An evaluation toolkit for simultaneous translation}.
\newblock In \emph{Proceedings of the 2020 Conference on Empirical Methods in
  Natural Language Processing: System Demonstrations}, pages 144--150, Online.
  Association for Computational Linguistics.

\bibitem[{Ma et~al.(2020{\natexlab{b}})Ma, Pino, and
  Koehn}]{ma-etal-2020-simulmt}
Xutai Ma, Juan Pino, and Philipp Koehn. 2020{\natexlab{b}}.
\newblock \href {https://doi.org/10.18653/v1/2020.aacl-main.58} {{S}imul{MT} to
  {S}imul{ST}: Adapting simultaneous text translation to end-to-end
  simultaneous speech translation}.
\newblock In \emph{Proceedings of the 1st Conference of the Asia-Pacific
  Chapter of the Association for Computational Linguistics and the 10th
  International Joint Conference on Natural Language Processing}, pages
  582--587, Suzhou, China. Association for Computational Linguistics.

\bibitem[{Matusov et~al.(2005)Matusov, Leusch, Bender, and
  Ney}]{matusov-etal-2005-evaluating}
Evgeny Matusov, Gregor Leusch, Oliver Bender, and Hermann Ney. 2005.
\newblock \href {https://aclanthology.org/2005.iwslt-1.19/} {Evaluating machine
  translation output with automatic sentence segmentation}.
\newblock In \emph{Proceedings of the Second International Workshop on Spoken
  Language Translation}, Pittsburgh, Pennsylvania, USA.

\bibitem[{McAuliffe et~al.(2017)McAuliffe, Socolof, Mihuc, Wagner, and
  Sonderegger}]{mcauliffe17_interspeech}
Michael McAuliffe, Michaela Socolof, Sarah Mihuc, Michael Wagner, and Morgan
  Sonderegger. 2017.
\newblock \href {https://doi.org/10.21437/Interspeech.2017-1386} {Montreal
  forced aligner: Trainable text-speech alignment using kaldi}.
\newblock In \emph{Interspeech 2017}, pages 498--502.

\bibitem[{Ouyang et~al.(2024)Ouyang, Xu, Dandekar, and
  Li}]{ouyang2024fasstfastllmbasedsimultaneous}
Siqi Ouyang, Xi~Xu, Chinmay Dandekar, and Lei Li. 2024.
\newblock \href {https://arxiv.org/abs/2408.09430} {Fasst: Fast llm-based
  simultaneous speech translation}.

\bibitem[{Papi et~al.(2024{\natexlab{a}})Papi, Gaido, Negri, and
  Bentivogli}]{papi-etal-2024-streamatt}
Sara Papi, Marco Gaido, Matteo Negri, and Luisa Bentivogli. 2024{\natexlab{a}}.
\newblock \href {https://doi.org/10.18653/v1/2024.acl-long.202} {{S}tream{A}tt:
  Direct streaming speech-to-text translation with attention-based audio
  history selection}.
\newblock In \emph{Proceedings of the 62nd Annual Meeting of the Association
  for Computational Linguistics (Volume 1: Long Papers)}, pages 3692--3707,
  Bangkok, Thailand. Association for Computational Linguistics.

\bibitem[{Papi et~al.(2022)Papi, Gaido, Negri, and
  Turchi}]{papi-etal-2022-generation}
Sara Papi, Marco Gaido, Matteo Negri, and Marco Turchi. 2022.
\newblock \href {https://doi.org/10.18653/v1/2022.autosimtrans-1.2}
  {Over-generation cannot be rewarded: Length-adaptive average lagging for
  simultaneous speech translation}.
\newblock In \emph{Proceedings of the Third Workshop on Automatic Simultaneous
  Translation}, pages 12--17, Online. Association for Computational
  Linguistics.

\bibitem[{Papi et~al.(2024{\natexlab{b}})Papi, Polak, Bojar, and
  Macháček}]{papi2024realrealtimesimultaneousspeechtotext}
Sara Papi, Peter Polak, Ondřej Bojar, and Dominik Macháček.
  2024{\natexlab{b}}.
\newblock \href {https://arxiv.org/abs/2412.18495} {How "real" is your
  real-time simultaneous speech-to-text translation system?}

\bibitem[{Papi et~al.(2023)Papi, Turchi, and Negri}]{alignatt}
Sara Papi, Marco Turchi, and Matteo Negri. 2023.
\newblock \href {https://doi.org/10.21437/Interspeech.2023-170} {Alignatt:
  Using attention-based audio-translation alignments as a guide for
  simultaneous speech translation}.
\newblock In \emph{Interspeech 2023}, pages 3974--3978.

\bibitem[{Post(2018)}]{post-2018-call}
Matt Post. 2018.
\newblock \href {https://doi.org/10.18653/v1/W18-6319} {A call for clarity in
  reporting {BLEU} scores}.
\newblock In \emph{Proceedings of the Third Conference on Machine Translation:
  Research Papers}, pages 186--191, Brussels, Belgium. Association for
  Computational Linguistics.

\bibitem[{Press et~al.(2021)Press, Smith, and Lewis}]{Press2021TrainST}
Ofir Press, Noah~A. Smith, and Mike Lewis. 2021.
\newblock \href {https://api.semanticscholar.org/CorpusID:237347130} {Train
  short, test long: Attention with linear biases enables input length
  extrapolation}.
\newblock \emph{ArXiv}, abs/2108.12409.

\bibitem[{Raffel et~al.(2024)Raffel, Agostinelli, and
  Chen}]{raffel-etal-2024-simultaneous}
Matthew Raffel, Victor Agostinelli, and Lizhong Chen. 2024.
\newblock \href {https://doi.org/10.18653/v1/2024.emnlp-main.1017}
  {Simultaneous masking, not prompting optimization: A paradigm shift in
  fine-tuning {LLM}s for simultaneous translation}.
\newblock In \emph{Proceedings of the 2024 Conference on Empirical Methods in
  Natural Language Processing}, pages 18302--18314, Miami, Florida, USA.
  Association for Computational Linguistics.

\bibitem[{Rei et~al.(2022)Rei, Treviso, Guerreiro, Zerva, Farinha, Maroti,
  C.~de Souza, Glushkova, Alves, Coheur, Lavie, and
  Martins}]{rei-etal-2022-cometkiwi}
Ricardo Rei, Marcos Treviso, Nuno~M. Guerreiro, Chrysoula Zerva, Ana~C Farinha,
  Christine Maroti, Jos{\'e}~G. C.~de Souza, Taisiya Glushkova, Duarte Alves,
  Luisa Coheur, Alon Lavie, and Andr{\'e} F.~T. Martins. 2022.
\newblock \href {https://aclanthology.org/2022.wmt-1.60/} {{C}omet{K}iwi:
  {IST}-unbabel 2022 submission for the quality estimation shared task}.
\newblock In \emph{Proceedings of the Seventh Conference on Machine Translation
  (WMT)}, pages 634--645, Abu Dhabi, United Arab Emirates (Hybrid). Association
  for Computational Linguistics.

\bibitem[{Ren et~al.(2020)Ren, Liu, Tan, Zhang, Qin, Zhao, and
  Liu}]{ren-etal-2020-simulspeech}
Yi~Ren, Jinglin Liu, Xu~Tan, Chen Zhang, Tao Qin, Zhou Zhao, and Tie-Yan Liu.
  2020.
\newblock \href {https://doi.org/10.18653/v1/2020.acl-main.350}
  {{S}imul{S}peech: End-to-end simultaneous speech to text translation}.
\newblock In \emph{Proceedings of the 58th Annual Meeting of the Association
  for Computational Linguistics}, pages 3787--3796, Online. Association for
  Computational Linguistics.

\bibitem[{Schneider and Waibel(2020)}]{schneider-waibel-2020-towards}
Felix Schneider and Alexander Waibel. 2020.
\newblock \href {https://doi.org/10.18653/v1/2020.iwslt-1.28} {Towards stream
  translation: Adaptive computation time for simultaneous machine translation}.
\newblock In \emph{Proceedings of the 17th International Conference on Spoken
  Language Translation}, pages 228--236, Online. Association for Computational
  Linguistics.

\bibitem[{Su(2023)}]{rerope2023}
Jianlin Su. 2023.
\newblock Rectified rotary position embeddings.
\newblock \url{https://github.com/bojone/rerope}.

\bibitem[{Su et~al.(2024)Su, Ahmed, Lu, Pan, Bo, and Liu}]{rope}
Jianlin Su, Murtadha Ahmed, Yu~Lu, Shengfeng Pan, Wen Bo, and Yunfeng Liu.
  2024.
\newblock \href {https://doi.org/10.1016/j.neucom.2023.127063} {Roformer:
  Enhanced transformer with rotary position embedding}.
\newblock \emph{Neurocomput.}, 568(C).

\bibitem[{Su et~al.(2021)Su, Lu, Pan, Wen, and Liu}]{Su2021RoFormerET}
Jianlin Su, Yu~Lu, Shengfeng Pan, Bo~Wen, and Yunfeng Liu. 2021.
\newblock \href {https://api.semanticscholar.org/CorpusID:233307138} {Roformer:
  Enhanced transformer with rotary position embedding}.
\newblock \emph{ArXiv}, abs/2104.09864.

\bibitem[{Sun et~al.(2023)Sun, Dong, Patra, Ma, Huang, Benhaim, Chaudhary,
  Song, and Wei}]{sun-etal-2023-length}
Yutao Sun, Li~Dong, Barun Patra, Shuming Ma, Shaohan Huang, Alon Benhaim,
  Vishrav Chaudhary, Xia Song, and Furu Wei. 2023.
\newblock \href {https://doi.org/10.18653/v1/2023.acl-long.816} {A
  length-extrapolatable transformer}.
\newblock In \emph{Proceedings of the 61st Annual Meeting of the Association
  for Computational Linguistics (Volume 1: Long Papers)}, pages 14590--14604,
  Toronto, Canada. Association for Computational Linguistics.

\bibitem[{Wang et~al.(2024)Wang, Vu, Wang, Shareghi, and
  Haffari}]{conversationsimulmt}
Minghan Wang, Thuy-Trang Vu, Yuxia Wang, Ehsan Shareghi, and Gholamreza
  Haffari. 2024.
\newblock \href {https://arxiv.org/abs/2402.10552} {Conversational simulmt:
  Efficient simultaneous translation with large language models}.

\bibitem[{Xiao et~al.(2024)Xiao, Tian, Chen, Han, and
  Lewis}]{xiao2024efficientstreaminglanguagemodels}
Guangxuan Xiao, Yuandong Tian, Beidi Chen, Song Han, and Mike Lewis. 2024.
\newblock \href {https://arxiv.org/abs/2309.17453} {Efficient streaming
  language models with attention sinks}.

\bibitem[{Xu et~al.(2024)Xu, Ouyang, Yan, Fernandes, Chen, Li, Neubig, and
  Watanabe}]{xu-etal-2024-cmus}
Xi~Xu, Siqi Ouyang, Brian Yan, Patrick Fernandes, William Chen, Lei Li, Graham
  Neubig, and Shinji Watanabe. 2024.
\newblock \href {https://doi.org/10.18653/v1/2024.iwslt-1.20} {{CMU}`s {IWSLT}
  2024 simultaneous speech translation system}.
\newblock In \emph{Proceedings of the 21st International Conference on Spoken
  Language Translation (IWSLT 2024)}, pages 154--159, Bangkok, Thailand
  (in-person and online). Association for Computational Linguistics.

\bibitem[{Yoshimura et~al.(2020)Yoshimura, Hayashi, Takeda, and
  Watanabe}]{yoshimura2020endtoendautomaticspeechrecognition}
Takenori Yoshimura, Tomoki Hayashi, Kazuya Takeda, and Shinji Watanabe. 2020.
\newblock \href {https://arxiv.org/abs/2002.00551} {End-to-end automatic speech
  recognition integrated with ctc-based voice activity detection}.

\bibitem[{Yu et~al.(2025)Yu, Zhao, Zhu, Xu, Zhou, and Zong}]{yu2025simulpl}
Donglei Yu, Yang Zhao, Jie Zhu, Yangyifan Xu, Yu~Zhou, and Chengqing Zong.
  2025.
\newblock \href {https://openreview.net/forum?id=XBF63bHDZw} {Simul{PL}:
  Aligning human preferences in simultaneous machine translation}.
\newblock In \emph{The Thirteenth International Conference on Learning
  Representations}.

\bibitem[{Zeng et~al.(2021)Zeng, Li, and Liu}]{zeng-etal-2021-realtrans}
Xingshan Zeng, Liangyou Li, and Qun Liu. 2021.
\newblock \href {https://doi.org/10.18653/v1/2021.findings-acl.218}
  {{R}eal{T}ran{S}: End-to-end simultaneous speech translation with
  convolutional weighted-shrinking transformer}.
\newblock In \emph{Findings of the Association for Computational Linguistics:
  ACL-IJCNLP 2021}, pages 2461--2474, Online. Association for Computational
  Linguistics.

\bibitem[{Zhang et~al.(2023)Zhang, Si, Chen, Zhang, Yang, Qu, and Jiao}]{lst}
Hao Zhang, Nianwen Si, Yaqi Chen, Wenlin Zhang, Xukui Yang, Dan Qu, and Xiaolin
  Jiao. 2023.
\newblock \href {https://arxiv.org/abs/2310.02050} {Tuning large language model
  for end-to-end speech translation}.

\end{thebibliography}

\appendix
\newpage
\section{Additional Data Details}
\label{sec:appdx_data}

\begin{figure}[t]
    \centering
    \includegraphics[width=0.8\linewidth]{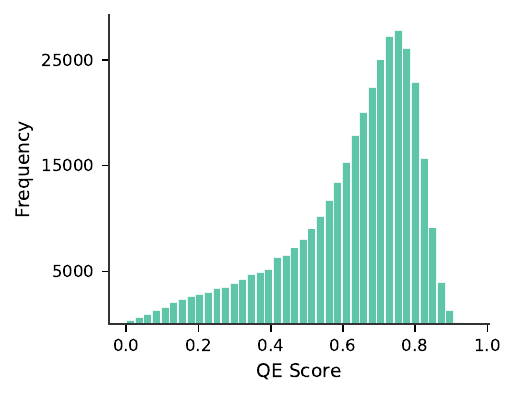}
    \caption{COMET-KIWI quality estimation score distribution on MuST-C en-zh data}
    \label{fig:comet_distribution}
\end{figure}

\begin{figure}[t]
    \centering
    \includegraphics[width=0.8\linewidth]{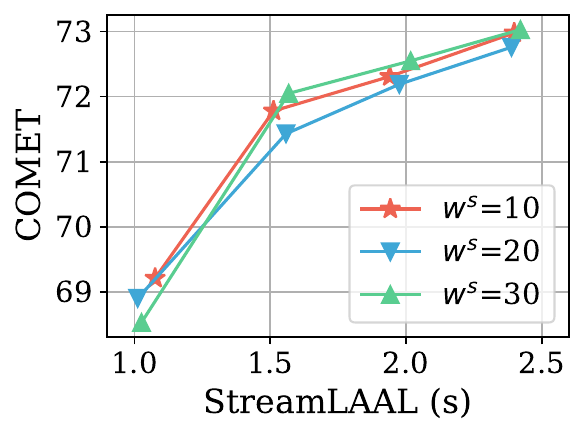}
    \caption{Impact of training-time sliding window size $w^s$ of speech encoder.}
    \label{fig:train_cache_speech}
\end{figure}

\begin{figure*}[t]
    \centering
    \includegraphics[width=\linewidth]{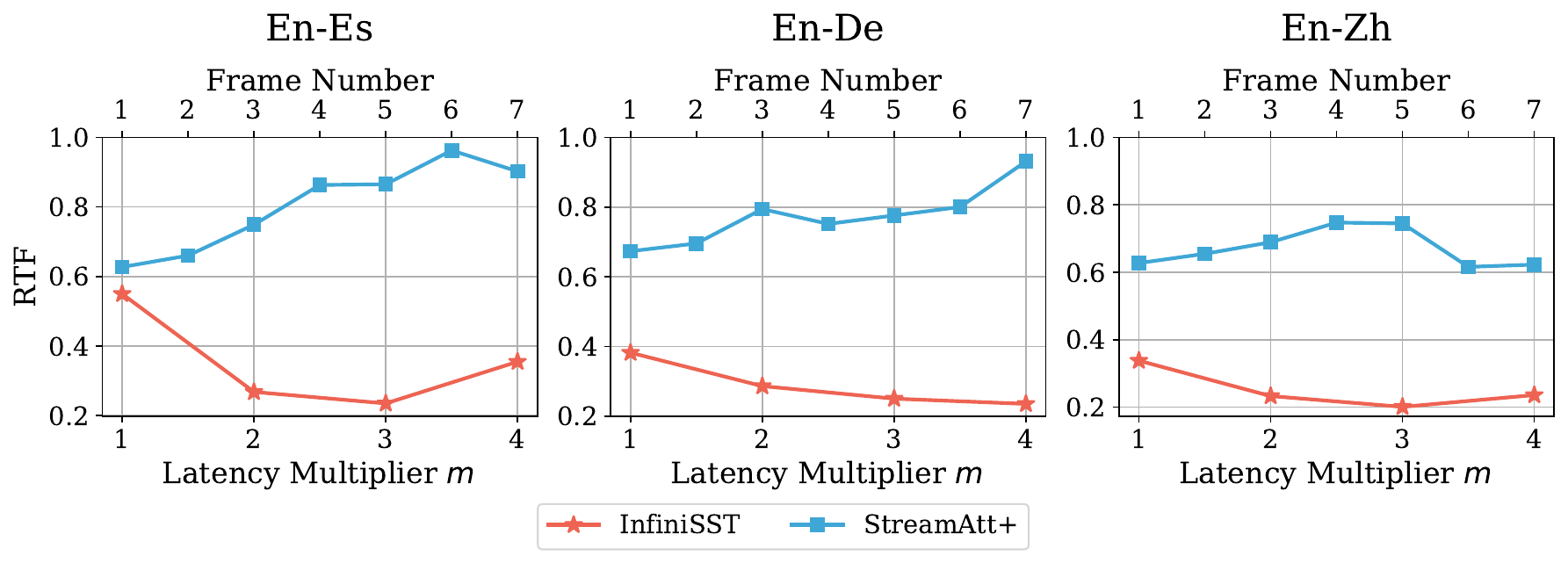}
    \caption{The Real-Time-Factor of \method~and baseline StreamAtt+.}
    \label{fig:rtf}
\end{figure*}

\begin{figure*}[t]
    \centering
    \includegraphics[width=\textwidth]{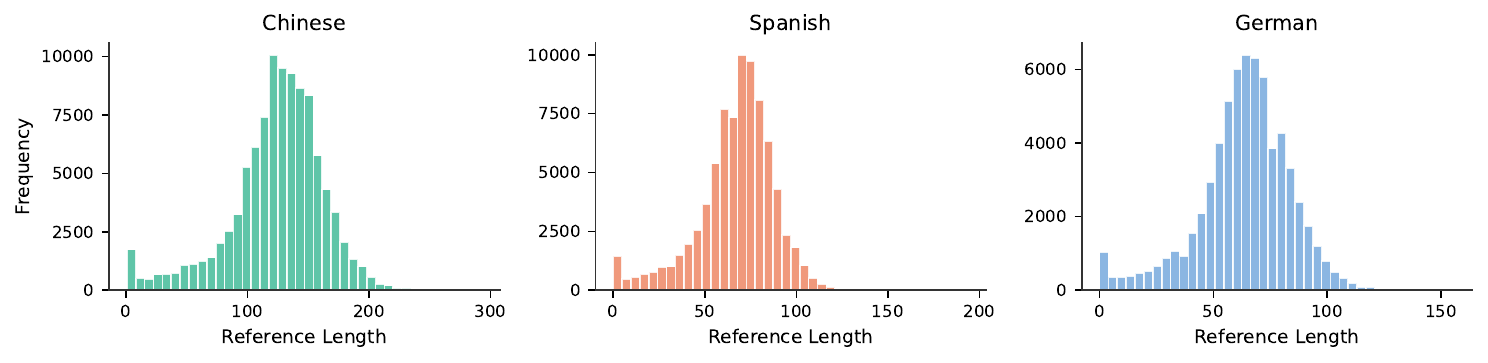}
    \caption{Reference length distribution of SST trajectories on MuST-C En-Zh, En-Es, and En-De.}
    \label{fig:reference_length_distribution}
\end{figure*}

\subsection{QE filtering and forward translation}
We first use Whisper to perform automatic speech recognition (ASR) on all training segments. We then apply CometKiwi\footnote{\url{https://huggingface.co/Unbabel/wmt23-cometkiwi-da-xxl}} to estimate the quality of ASR outputs by computing quality estimation (QE) scores between the ASR results and the reference text. As shown in Figure~\ref{fig:comet_distribution}, we retain only instances where the QE score is greater than 0.5, which accounts for 78.64\% of the data, resulting in a total of 280K instances.

Upon further inspection, we observed that many filtered-out cases exhibited acceptable word error rates (WER) between the ASR outputs and the source text. To recover these cases, we performed forward translation using the 7B version of TowerInstruct \footnote{\url{https://huggingface.co/Unbabel/TowerInstruct-7B-v0.2}}on the source text using TowerInstruct with the following decoding settings: temperature = 0.0 and frequency penalty = 0.1. The translations were generated using vLLM.

\subsection{Dataset Statistics}
The MuST-C dataset used in our experiments consists of 105,647 instances for En-Zh, 88,725 for En-Es, and 70,037 for En-De.

Figure~\ref{fig:reference_length_distribution} shows the reference length distribution across these language pairs.

For En-Zh, the reference text length averages 124.32 characters, with a maximum of 444. En-Es has significantly longer references, averaging 400.06 characters and reaching a maximum of 1,116. En-De also exhibits long references, with an average of 419.47 characters and a maximum of 957. Spanish reference lengths in word count average 67.1 words, with a median of 70.0 and a 90th percentile of 90.0. German references are slightly shorter, averaging 63.6 words, with a median of 65.0 and a 90th percentile of 87.0.

En-Zh segments average 26.85 seconds, En-Es 25.25 seconds, and En-De 26.11 seconds, all with a maximum of 28.80 seconds, reflecting speech-text alignment across languages.

\subsection{Examples of Trajectory}

\begin{figure*}[t]
    \centering
    \includegraphics[width=1.0\linewidth]{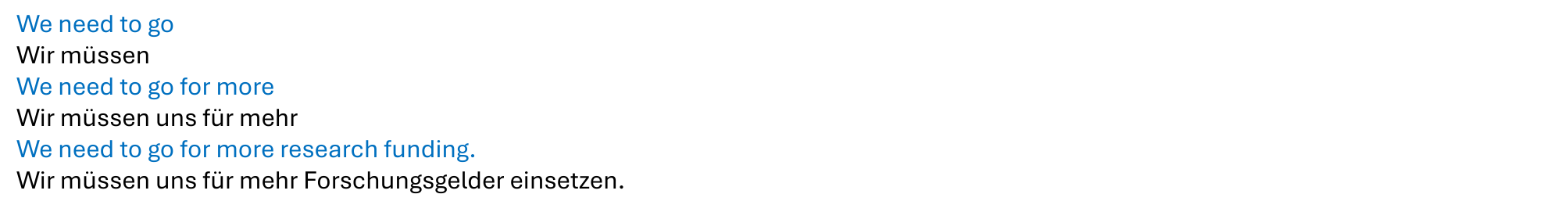}
        \includegraphics[width=1.0\linewidth]{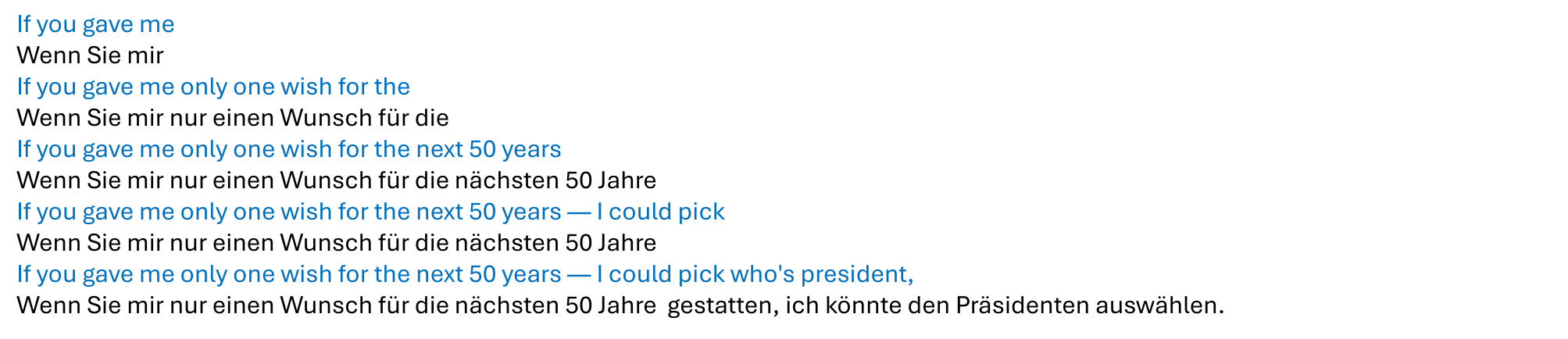}
    \includegraphics[width=1.0\linewidth]{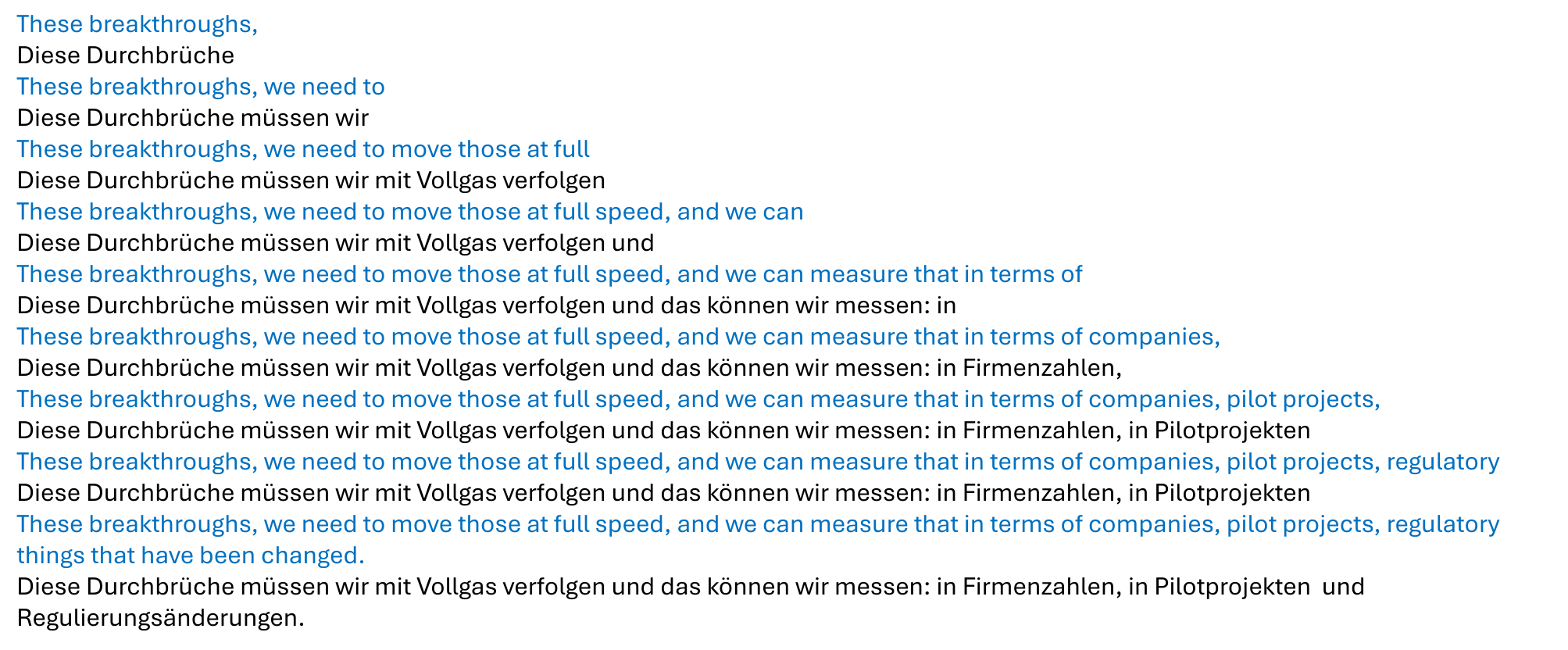}
    \caption{Three examples of En-De trajectory.}
    \label{fig:traj_eg}
\end{figure*}

We show additional examples of trajectory in Figure \ref{fig:traj_eg}.

\section{Additional Experiment Results}

Impact of speech encoder window size during training $w^s$ is shown in Figure \ref{fig:train_cache_speech}. The RTF of \method and baseline StreamAtt+ is shown in Figure \ref{fig:rtf}.

\end{document}